\documentclass[11pt]{article}

\usepackage{penglab-preprint}
\usepackage{booktabs}
\usepackage{graphicx}
\usepackage[numbers,comma,sort&compress]{natbib} 

\usepackage[inline]{enumitem}
\usepackage{colortbl}
\usepackage{wrapfig}
\usepackage{subcaption}
\usepackage{tabularx}
\usepackage{makecell}
\usepackage{placeins}
\usepackage{zref-clever}
\usepackage{float}
\usepackage{siunitx}

\definecolor{graylinecolor}{HTML}{C1C7D6}
\newcommand{\grayline}{%
    \arrayrulecolor{graylinecolor}%
    \hline%
    \arrayrulecolor{black}%
    \noalign{\vskip 2pt}%
}

\zcsetup{
    cap=true,
    nameinlink=false,
}

\zcLanguageSetup{english}{
type = etable,
    Name-sg = eTable,  
    name-sg = eTable,
    Name-pl = eTables, 
    name-pl = eTables,
type = efigure,
    Name-sg = eFigure,  
    name-sg = eFigure,
    Name-pl = eFigures, 
    name-pl = eFigures,
type = esection,
    Name-sg = eSection,  
    name-sg = eSection,
    Name-pl = eSections, 
    name-pl = eSections,
}

\newcolumntype{L}{>{\raggedright\arraybackslash}X}

\title{A Disease-specific Retinal Aging Biomarker for Predicting Progression to Late Age-related Macular Degeneration}

\author[1,$\dagger$]{Hexin Dong}
\author[2,$\dagger$]{Benjamin Hou}
\author[1]{Puzhen Wu}
\author[3]{Mingquan Lin}
\author[4]{Emily Y. Chew}
\author[2,*]{Zhiyong Lu}
\author[1,*]{Yifan~Peng}
\affil[1]{Population Health Sciences, Weill Cornell Medicine, New York, 10016, USA}
\affil[2]{National Library of Medicine, National Institutes of Health, Bethesda, 20894, USA}
\affil[3]{Department of Surgery, University of Minnesota, Minneapolis, 55455, USA}
\affil[4]{National Eye Institute, National Institutes of Health, Bethesda, 20894, USA}
\affil[*]{Corresponding author(s). Email(s): \url{zhiyong.lu@nih.gov}, \url{yip4002@med.cornell.edu}}
\affil[$\dagger$]{These authors contributed equally to this work.}

\begin{document}

\maketitle

\begin{abstract}
Age-related macular degeneration (AMD) is a leading cause of irreversible vision loss in older adults. While chronological age is a primary risk factor, it inadequately reflects individual variations in retinal aging, limiting effective AMD risk stratification and timely intervention. Here, we propose and validate fundus-based AMD age (FAA), which quantifies disease-specific retinal aging in AMD and captures accelerated AMD-related retinal aging, thereby improving prognostic precision over chronological age. Trained on 36,782 images from 2,192 participants in the Age-Related Eye Disease Study (AREDS) dataset and validated across AREDS and AREDS2 cohorts, our model demonstrates significant improvements over traditional chronological age-based approaches. Specifically, integrating the proposed FAA into Cox proportional-hazards models significantly enhances prediction performance. FAA improved the C-index from 0.836 to 0.888 and reduced the 5-year Brier score from 0.050 to 0.046 in AREDS, with consistent gains observed in AREDS2 (C-index 0.715 to 0.756; 2-year Brier 0.078 to 0.076). FAA showed improved prognostic performance across the two datasets and under different phenotype definitions (reading center--defined vs. specialist-defined grading), supporting its potential utility for AMD risk stratification. These findings highlight the advantage of incorporating FAA into AMD risk prediction frameworks, supporting further evaluation of FAA for AMD risk stratification. 
\end{abstract}


\section{Introduction}

Age-related macular degeneration (AMD) is one of the leading causes of irreversible vision impairment and blindness among elderly individuals worldwide, with late-stage AMD dramatically impacting quality of life due to central vision loss~\cite{ferris2013clinical,fleckenstein2024age,paudel2024economic,keenan2021age}. The progression to late AMD significantly reduces visual function, independence, and overall well-being, underscoring the urgent need for early detection and timely intervention~\cite{ambati2003age,apte2021age,zhang2024global,mitchell2018age,chuck2021comprehensive}.

Historically, chronological age, defined as the amount of time that has passed since a person's birth, has played a central role in guiding clinical decisions related to prevention, screening, and intervention. In the context of AMD progression, chronological age is also widely recognized as an important risk factor for predicting AMD, as it is closely associated with the structural and functional decline of the eye. Additionally, clinicians utilize imaging biomarkers from color fundus photography (CFP)~\cite{age2001age} and combine them with demographic information, such as age, for diagnosis and prognosis~\cite{chuck2021comprehensive,age2001age,kanagasingam2014progress,grassmann2018deep}. However, the current state-of-the-art AMD risk calculator, which relies on chronological age, presents limitations~\cite{chuck2021comprehensive,chiu2014risk,sitnilska2020major}. There is substantial interindividual variation in AMD progression among individuals of the same chronological age, suggesting that chronological age does not fully capture retinal characteristics associated with AMD progression~\cite{lowsky2014heterogeneity,seddon2011risk}. This discrepancy motivates the development of an AMD-related retinal age that can complement chronological age in prognostic modeling, offer greater prognostic precision, and help tailor early interventions to effectively prevent or delay progression to the late stage of AMD.

Recently, deep learning methods have shown promise for estimating biological aging clocks using regression models trained on retinal images~\cite{yu2023retinal,miao2025artificial,li2025large}. Typically, these approaches are developed using fundus images from healthy individuals and then use either the estimated biological age or the age gap (the difference between machine-learning-predicted age and chronological age) as biomarkers for disease risk prediction~\cite{yu2023retinal,liu2019biological,zhu2023retinal,ahadi2023longitudinal}. However, these approaches do not capture disease-specific image-based aging processes and may not fully capture disease-related retinal variation relevant to AMD progression, where retinal structure and progression dynamics differ substantially from healthy eyes.

Deep learning has also been applied directly to AMD prognosis from color fundus photographs. Prior models have predicted progression to late AMD using baseline CFPs, clinical or genetic covariates, survival-analysis frameworks, and longitudinal image sequences~\cite{holste2024harnessing,yan2020deep,peng2020predicting,gervelmeyer2024interpretable,lee2022predicting}. For example, a prior longitudinal CFP study used CNN--LSTM and CNN--Transformer models to capture sequential information from patients' image histories and predict 2-year and 5-year risk of progression to late AMD~\cite{lee2022predicting}. These approaches demonstrate the prognostic value of retinal imaging, but they primarily produce direct image-to-risk estimation rather than an interpretable disease-specific aging biomarker that can be incorporated into conventional survival models. Therefore, it remains unclear whether fundus-derived retinal aging signals can improve AMD risk prediction beyond chronological age, retinal age gap, and conventional AMD severity grading.

To address these gaps, we introduce \textbf{F}undus-based \textbf{A}MD \textbf{A}ge (\textbf{FAA}), an AI-derived disease-specific retinal aging biomarker that quantifies accelerated fundus aging associated with AMD progression. Unlike epigenetics-based clocks developed for general aging and mortality estimation, FAA captures AMD-related retinal aging signatures learned directly from CFP annotated with validated AMD severity scores. To our knowledge, this represents the first effort utilizing a fundus-derived age within survival analysis frameworks explicitly tailored to AMD prognosis. Moreover, our method offers greater interpretability than existing deep learning-based AMD prognosis models~\cite{holste2024harnessing,yan2020deep,peng2020predicting,gervelmeyer2024interpretable,lee2022predicting}, which typically predict risk from CFP in a black-box manner. By explicitly predicting fundus-based AMD age, our model enables a more accurate understanding and communication of individual risk estimates, which are crucial for clinical decision-making.

Our method was developed using two independent datasets from the Age-Related Eye Disease Studies (AREDS~\cite{study1999age}and AREDS2~\cite{chew2012age}), collected by the National Eye Institute in the U.S., which have been extensively used in research on AMD prognosis. We trained the FAA estimation model on 36,782 retinal images from 2,192 participants in the AREDS. We then combined the FAA with biomarkers defined by either a human grader at the reading center or retinal specialists to construct Cox proportional hazards models for AMD progression. The prognostic models were evaluated using phenotype-specific eye-level samples. The reading-center-defined analyses included 769 eyes from the AREDS internal-test cohort and 5,651 eyes from AREDS2, whereas the specialist-defined analyses included 811 and 5,964 eyes, respectively.

\section{Results}\label{results}

\subsection{Overview}\label{overview}

\begin{figure}[t!]
\centering
\includegraphics[width=\linewidth]{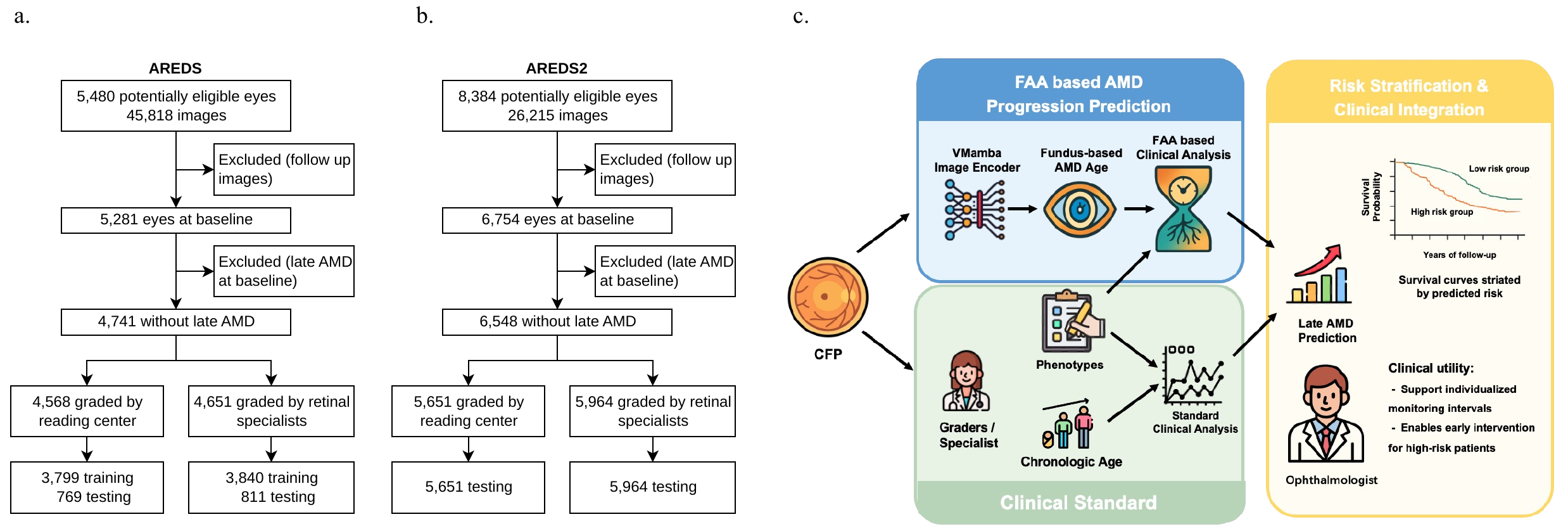}
\caption{\textbf{Overview of the study workflow for predicting late AMD progression.} \textbf{a-b}, Flowchart detailing participant selection, data preprocessing, and exclusion criteria for building the AMD risk prediction models based on the AREDS and AREDS2 datasets. \textbf{c}, The workflow for predicting late AMD progression. FAA, as predicted by the VMamba model~\cite{liu2024vmamba} from CFP, is combined with smoking status and clinically relevant AMD phenotypes (e.g., drusen size and pigmentary abnormalities), and the resulting model is integrated into a Cox proportional hazards model to predict the risk of AMD progression.}
\label{fig:overview}
\end{figure}

We conducted a series of experiments using the data from the Age-Related Eye Disease Studies (AREDS~\cite{study1999age} and AREDS2~\cite{chew2012age}), the largest longitudinal clinical trials in AMD (\zcref{fig:overview}a and \zcref{fig:overview}b, detailed in \textbf{Methods}). The AREDS dataset comprises 45,818 fundus images from 2,740 participants. Dataset allocation was performed at the participant level, with both eyes and all longitudinal visits from the same participant retained within a single partition. After applying baseline endpoint, follow-up, and phenotype-availability criteria, the reading-center-defined Cox analyses included 3,799 development eyes and 769 internal-test eyes. These eyes were graded by certified human graders at the University of Wisconsin Reading Center (hereinafter referred to as the \textbf{reading-center-defined setting}). To compare the performance of FAA with that of retinal specialists (investigators whose assessments had previously been recorded during AREDS), we also constructed a \textbf{specialist-defined} setting graded by retinal specialists at the qualifying visit, using a subset of 3,840 eyes from the training set and 811 eyes from the test set.

AREDS2 includes 26,215 fundus images from 4,192 participants and was used as the external testing set for this study. After phenotype-specific completeness criteria and prespecified participant-level exclusions were applied, the final external-validation samples included 5,651 eyes in the reading-center-defined analysis and 5,964 eyes in the specialist-defined analysis. No AREDS2 data were used for FAA training, Cox-model fitting, model selection, hyperparameter optimization, or threshold determination.

\zcref{fig:overview}c shows the overall framework of late AMD progression prediction. First, a VMamba model was adapted to estimate FAA~\cite{liu2024vmamba}. The model was trained on all CFPs from the AREDS training set to ensure generalizability across the full spectrum of AMD severity (see described under FAA model development for training details). Second, a Cox proportional hazards model was used to predict the probability of progression to late AMD based on FAA. In this step, we added additional phenotypes to further enhance the model's predictive performance.

The specific aims of this study were to evaluate the prognostic contribution of FAA when combined with phenotypes assessed by certified reading-center graders or study retinal specialists. For different scenarios, we utilize distinct phenotypes to ensure a fair comparison. The specific details and rationale for selecting these phenotypes are provided in \zcref{sec:dataset-description}.

\subsection{FAA improves AMD progression prediction with phenotypes graded by the Reading Center}\label{faa-improves-amd-progression-prediction-with-phenotypes-graded-by-the-reading-center}

\begin{figure}[t!]
\centering
\includegraphics[width=1.0\linewidth]{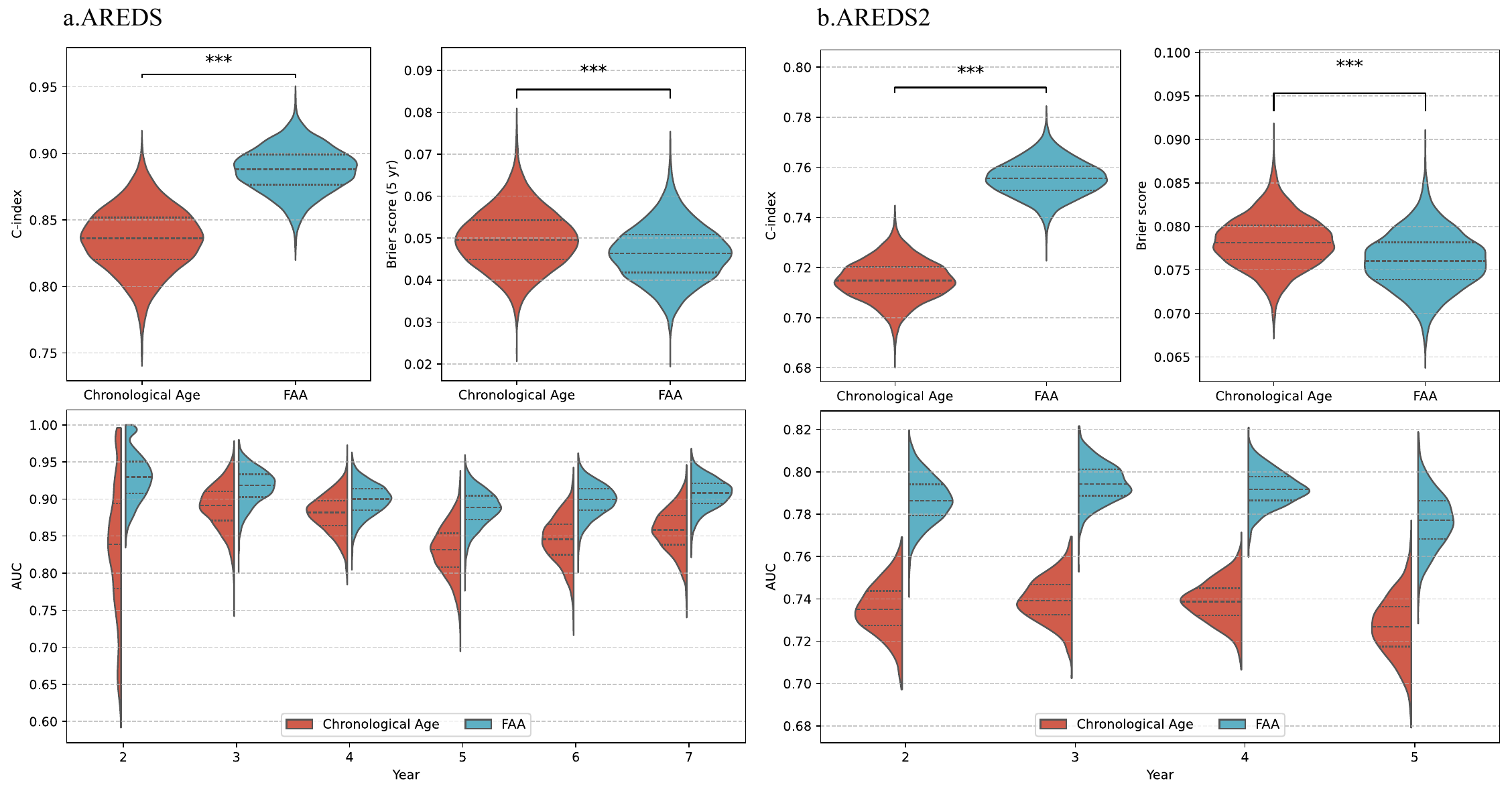}
\caption{\textbf{Comparison of prognostic performance between models using age (chronological age or FAA) and phenotypes graded by reading centers in AREDS and AREDS2.} \textbf{a}, AREDS. \textbf{b}, AREDS2. P-values were calculated using two-sided bootstrap resampling (5,000 iterations). Statistical significance was indicated as *** for $p<0.001$.}
\label{fig:prognostic performance}
\end{figure}

\zcref{fig:prognostic performance} compares the performance of models using age (chronological age or FAA) and the phenotypes graded by the \emph{Reading Center} (geographic atrophy within grid [GEOAWI], RPE depigmentation area within grid [RPEDWI], increased pigment area within grid [INCPWI], and drusen area within grid [DRARWI]). \zcref{fig:prognostic performance}a shows that the model with FAA exhibited a significant improvement in prediction accuracy compared to the one with chronological age on AREDS, reflected by a higher C-index (0.888 [95\% CI:0.853-0.919] vs. 0.836 [95\% CI: 0.788-0.880], $p<0.001$) and lower Brier scores at 5 years (0.046 [95\% CI:0.033-0.060 ] vs. 0.050 [95\% CI: 0.037-0.063], $p<0.001$). Moreover, time-dependent AUC analyses showed that the FAA-based model achieved higher AUC point estimates than the chronological-age-based model at every evaluated horizon. Complete year-specific results and 95\% confidence intervals are provided in \zcref{ssec:year-specific, tab:supp-time-dependent-auc}.

Similar performance improvements were observed in the external AREDS2 dataset (\zcref{fig:prognostic performance}b). The model with FAA significantly improved the C-index from 0.715 [95\% CI: 0.699-0.730] to 0.756 [95\% CI: 0.742-0.770] ($p<0.001$). The FAA-based model also produced a reduction in the 2-year Brier score from 0.078 [95\% CI: 0.073--0.084] to 0.076 [95\% CI: 0.070-0.082] ($p<0.001$). Time-dependent AUC results also demonstrated consistently superior performance of the model using FAA compared to chronological age throughout years 2 to 5 (Detailed results in \zcref{ssec:year-specific}).

\subsection{FAA improves AMD progression prediction with phenotypes graded by retinal specialists}\label{faa-improves-amd-progression-prediction-with-phenotypes-graded-by-retinal-specialists}

\begin{figure}[t!]
\centering
\includegraphics[width=1.0\linewidth]{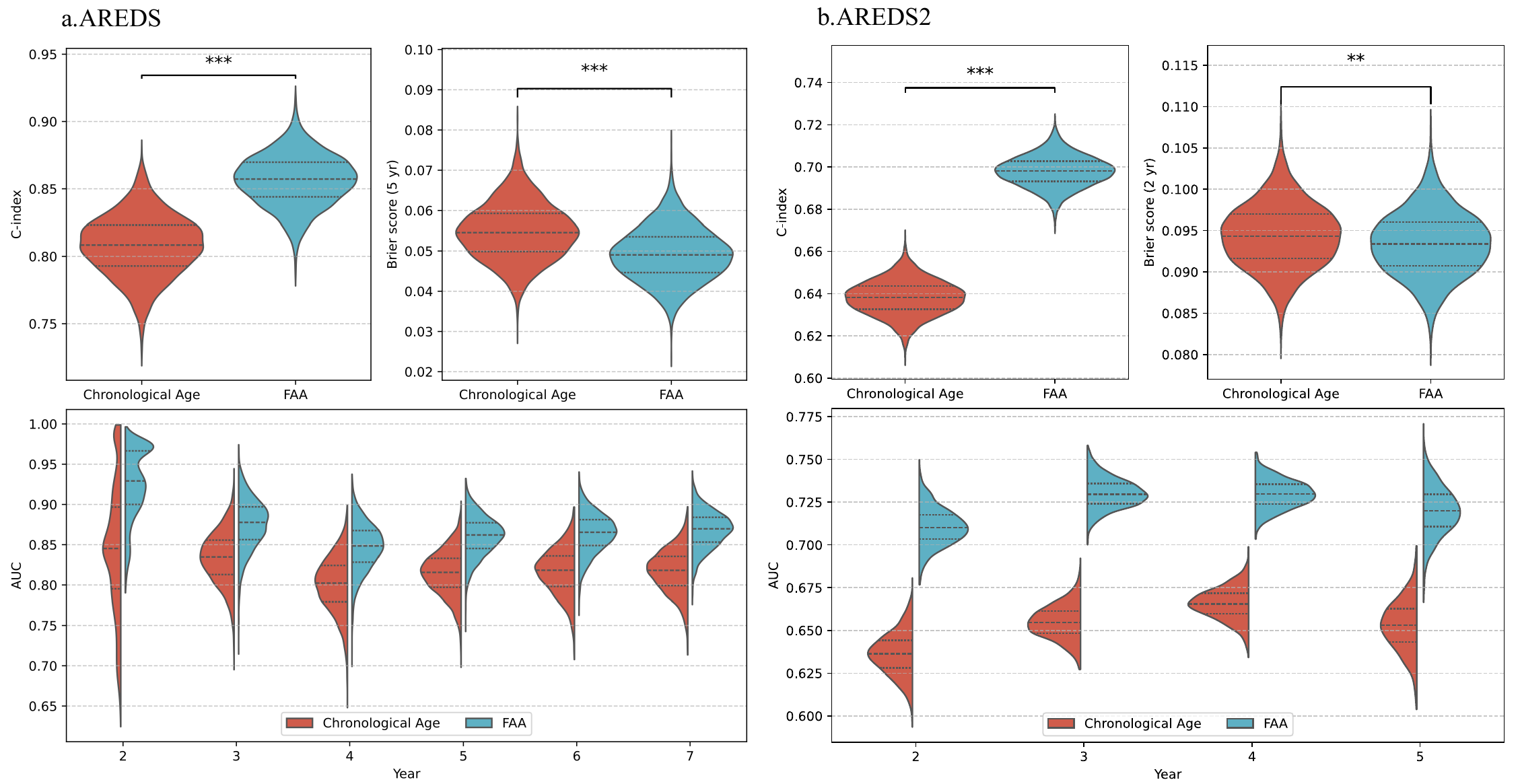}
\caption{\textbf{Comparison of prognostic performance between models using age (chronological age or FAA) and phenotypes graded by human retinal specialists on AREDS and AREDS2.} \textbf{a}, AREDS. \textbf{b}, AREDS2. p-values were calculated using two-sided bootstrap resampling (5,000 iterations). Statistical significance was indicated as *** for $p<0.001$, ** for $p<0.01$.}
\label{fig:prognostic performance human}
\end{figure}

\zcref{fig:prognostic performance human} compares the performance of predictive models using age (chronological age or FAA) and phenotypes graded by \emph{retinal specialists} at the qualifying visit (clinical drusen size assessment [DRUSENQ], pigment abnormalities [PIGDRUQ]). \zcref{fig:prognostic performance human} shows that, despite the performance gap introduced by variations in phenotype usage, the improvements achieved by incorporating FAA remain consistent with those observed in \zcref{fig:prognostic performance}. On AREDS, using specialist-defined phenotypes, the model with FAA has a performance increase in C-index from 0.808 [95\% CI: 0.761-0.851] to 0.857 [95\% CI: 0.817-0.892] ($p<0.001$) and in Brier Score from 0.055 [95\% CI: 0.042-0.069] to 0.049 [95\% CI: 0.037-0.063] ($p<0.001$). In addition, the model with FAA exhibited consistently higher time-dependent AUCs across years 2 to 7, compared to the chronological age-based model (detailed results in \zcref{ssec:year-specific, tab:supp-time-dependent-auc-human}). The relative improvement in discrimination was observed under both phenotype-definition settings.

Similar performance improvements were observed in the external AREDS2 dataset (\zcref{fig:prognostic performance human}b). FAA-based model significantly improved the C-index from 0.638 [95\% CI: 0.622-0.653] to 0.698 [95\% CI: 0.683-0.712] ($p<0.001$) and reduced the 2-year Brier score from 0.094 [95\% CI: 0.087-0.102] to 0.093 [95\% CI: 0.086-0.101] ($p<0.01$). Time-dependent AUC results also demonstrated consistently superior performance of FAA compared to chronological age throughout years 2 to 5. Complete year-specific AUC results and 95\% confidence intervals are provided in \zcref{ssec:year-specific, tab:supp-time-dependent-auc-human}.

\subsection{Adjusted associations of FAA and chronological age with late-AMD progression}\label{adjusted-associations-of-faa-and-chronological-age-with-late-amd-progression}

To examine the associations of FAA and chronological age with late-AMD progression after adjustment for established clinical covariates, we fitted two otherwise identically specified Cox proportional hazards models within each phenotype-definition setting. Both models were fitted using the same analytical eyes and included the same smoking and AMD phenotype covariates; the only difference was whether chronological age or FAA was included as the age-related predictor. Both age representations were modeled per 1-year increase.

\begin{figure}[t!]
\centering
\includegraphics[width=1.0\linewidth]{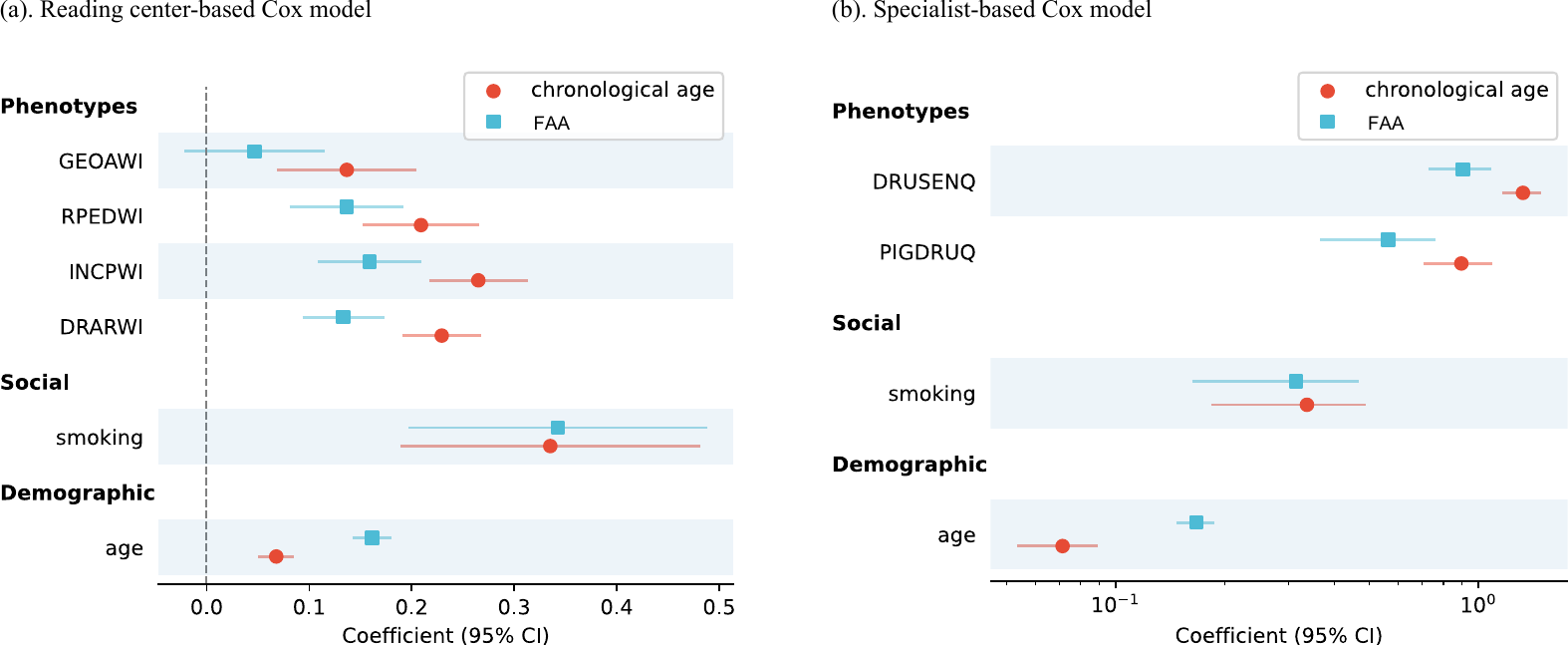}
\caption{\textbf{Multivariate Cox regression analyses adjusted for clinical covariates.} \textbf{a}, Reading-center-defined Cox regression model using chronological age versus FAA. Covariates included geographic atrophy within the grid (GEOAWI), RPE depigmentation area within the grid (RPEDWI), increased pigment area within the grid (INCPWI), drusen area within the grid (DRARWI), and smoking status (SMK). \textbf{b}, Specialist-defined Cox regression model using chronological age versus FAA. Covariates included clinical drusen size assessment (DRUSENQ) and pigment abnormalities (PIGDRUQ), alongside smoking status. Each row represents the coefficient and its corresponding 95\% confidence interval. Hazard ratios correspond to a one-unit increase in each covariate (1 year for FAA and chronological age, and one-grade increase for ordinal phenotype variables).}
\label{fig:cox}
\end{figure}

In the reading-center-defined setting (\zcref{fig:cox}a), each 1-year increase in FAA was associated with a 17.5\% higher hazard of late-AMD progression (HR, 1.175; 95\% CI, 1.153--1.198), whereas each 1-year increase in chronological age was associated with a 7.0\% higher hazard in the corresponding matched model (HR, 1.070; 95\% CI, 1.051--1.090). A similar pattern was observed in the specialist-defined setting (\zcref{fig:cox}b), in which the adjusted HRs per 1-year increase were 1.182 (95\% CI, 1.158--1.206) for FAA and 1.074 (95\% CI, 1.054--1.093) for chronological age.

These results indicate that FAA had a numerically larger per-year adjusted association with late-AMD progression than chronological age in otherwise matched models. However, because FAA and chronological age were evaluated in separately fitted models and FAA is a learned, disease-informed image representation, the coefficient magnitudes were not interpreted as a formal ranking of variable importance or as evidence that one FAA year is biologically equivalent to one calendar year. The comparative prognostic performance of the two age representations was instead evaluated using held-out C-index, time-dependent AUC, and Brier-score results, as reported in the held-out prognostic-performance analyses.

\subsection{Prognostic value of FAA in predicting AMD progression}\label{prognostic-value-of-faa-in-predicting-amd-progression}

\begin{figure}[t!]
\centering
\includegraphics[width=\linewidth]{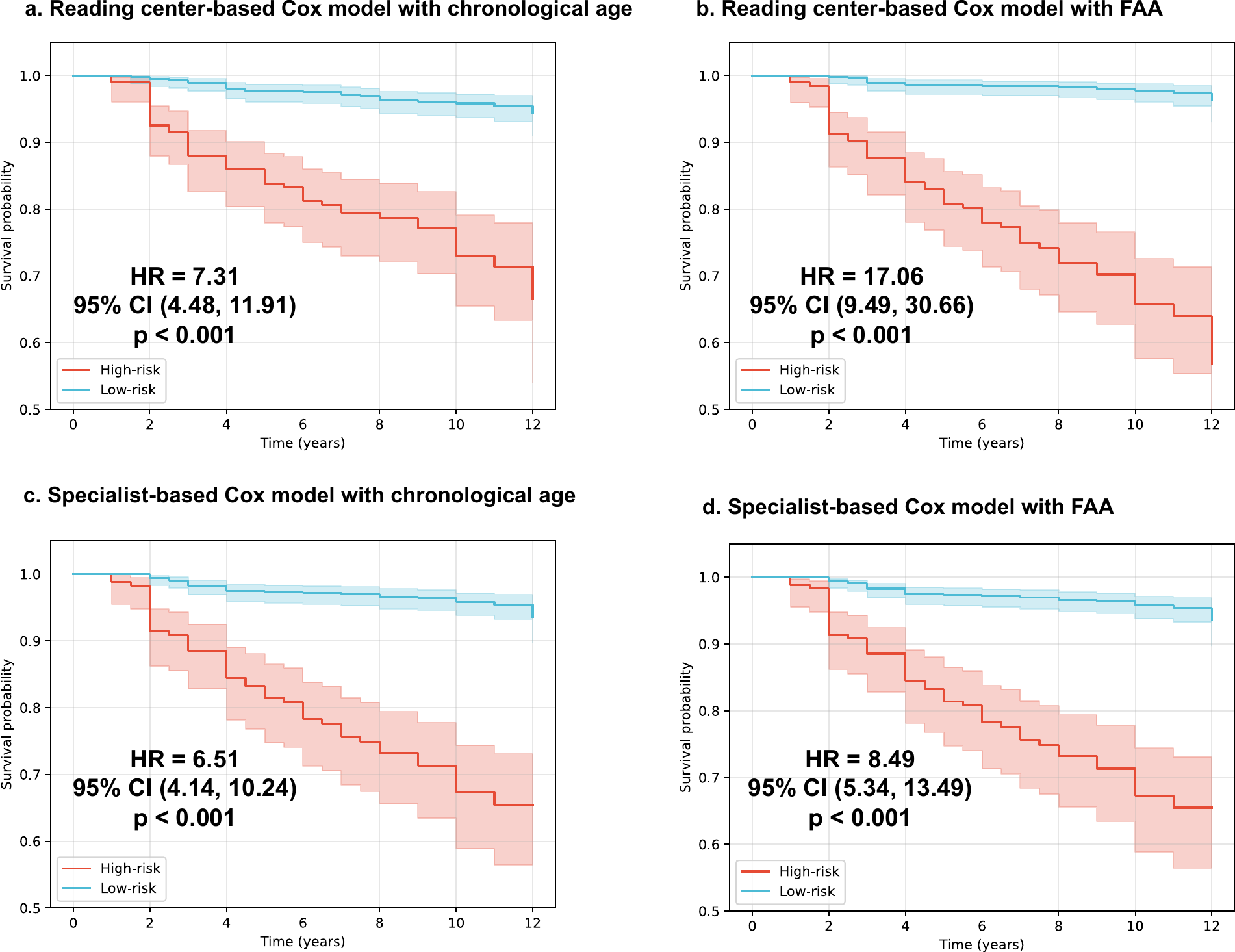}
\caption{\textbf{Kaplan-Meier survival curves comparing the prognostic performance of Cox proportional hazards models in predicting overall AMD progression (from early stages to late AMD) using chronological age and FAA.} \textbf{a-b}, Reading-center-defined Cox models using chronological age and FAA. \textbf{c-d}, Specialist-defined Cox models using chronological age and FAA. High- and low-predicted-risk groups were defined using the model-specific median predicted risk in the AREDS development cohort. The fixed thresholds were applied unchanged to the AREDS internal-test cohort.}
\label{fig:km}
\end{figure}

\zcref{fig:km} shows Kaplan-Meier (KM) survival curves comparing Cox proportional hazards models that use either chronological age or FAA to predict AMD progression from early stages (including no AMD, early AMD, and intermediate AMD) to late AMD. Eligible eyes were stratified into high-risk and low-risk groups based on the median risk score derived from the training dataset. Eligible eyes with risk scores exceeding the median were classified as high-risk, while those with scores at or below the median were considered low-risk. Both the reading-center-based (\zcref{fig:km}a-b) and specialist-defined (\zcref{fig:km}c-d) Cox models demonstrated greater separation in survival probabilities between high- and low-risk groups with FAA than with chronological age. This finding underscores the FAA's superior prognostic value as a discriminative biomarker for predicting AMD progression.

\begin{figure}[t!]
\centering
\includegraphics[width=1.0\linewidth]{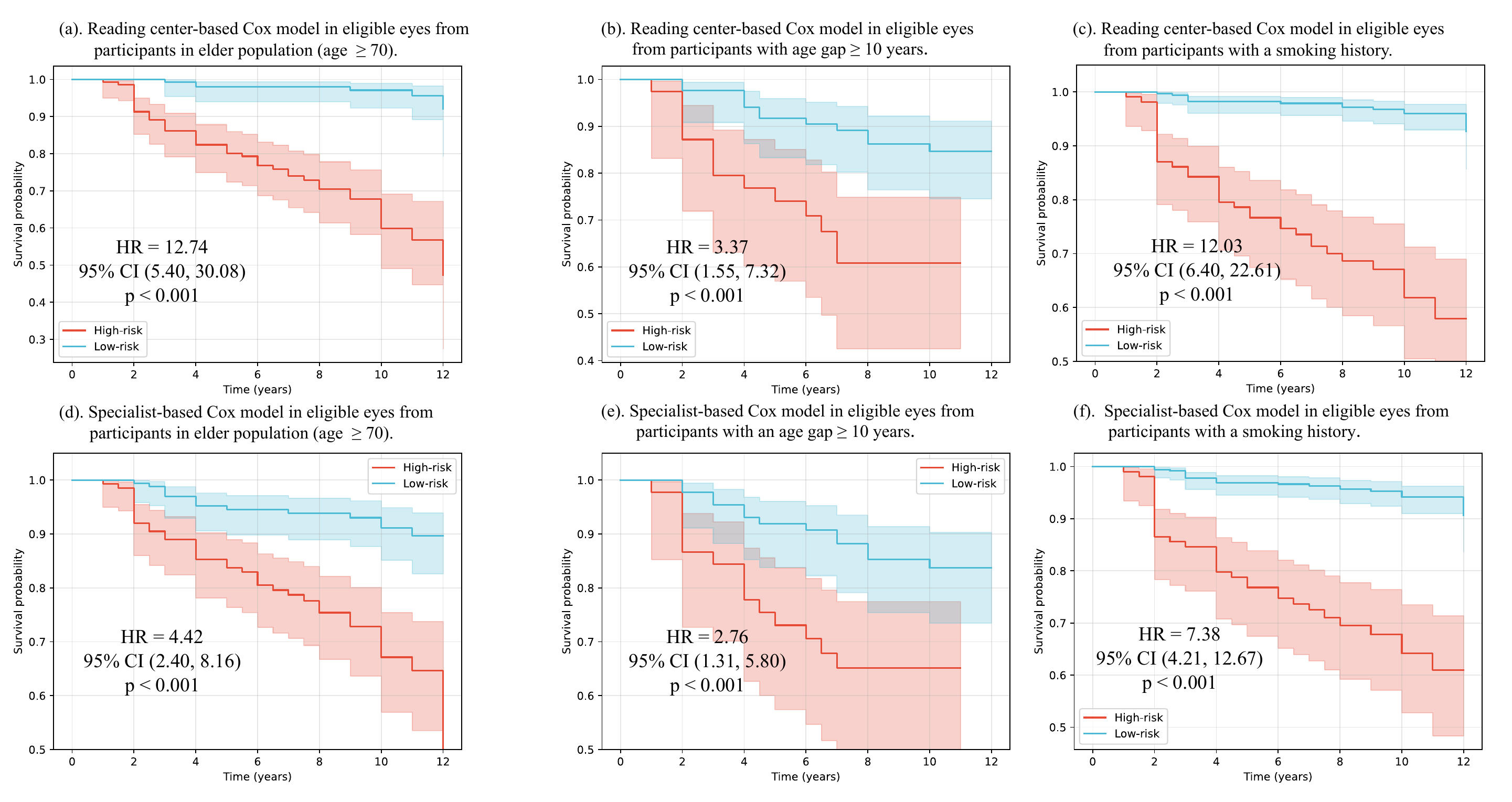}
\caption{\textbf{Kaplan--Meier survival analyses combine FAA with other clinical risk factors.} \textbf{a-c}, Reading-center-defined Cox models in (a) Eligible eyes from participants aged $\ge 70$ years, (b) Eligible eyes from participants with a fundus-based AMD age gap of $\ge 10$ years, and (c) Eligible eyes from participants with a smoking history. \textbf{d-f}, Specialist-defined Cox models in (d) Eligible eyes from participants aged $\ge 70$ years, (e) Eligible eyes from participants with a fundus-based AMD age gap of $\ge 10$ years, and (f) Eligible eyes from participants with a smoking history. Across all subgroups, integrating FAA into the Cox model identifies high-risk eyes, underscoring the added prognostic value of combining FAA with other clinical factors.}
\label{fig:km others}
\end{figure}

We further investigated whether integrating FAA with additional risk factors could enhance the stratification of eligible eyes at higher risk for AMD progression. \zcref{fig:km others} shows Kaplan-Meier (KM) survival analyses comparing the prognostic performance of reading-center-defined (\zcref{fig:km others}a-c) and specialist-defined (\zcref{fig:km others}d-f) Cox models across three clinically relevant subgroups: eligible eyes from participants aged $\ge 70$ years, eligible eyes from participants with $\ge 10$ years of age gap between FAA and chronological ages, and eligible eyes from participants with a smoking history. For each subgroup, the cutoff was defined as the median predicted risk among developing eyes satisfying the same subgroup criterion and was applied unchanged to the corresponding internal-test subgroup. These findings indicate that combining FAA and risk factors can more effectively identify eligible eyes at substantially increased risk of AMD progression.

\subsection{Characterization of FAA in relation to chronological age and AMD severity}

\begin{figure}[t!]
\centering
\includegraphics[width=1.0\linewidth]{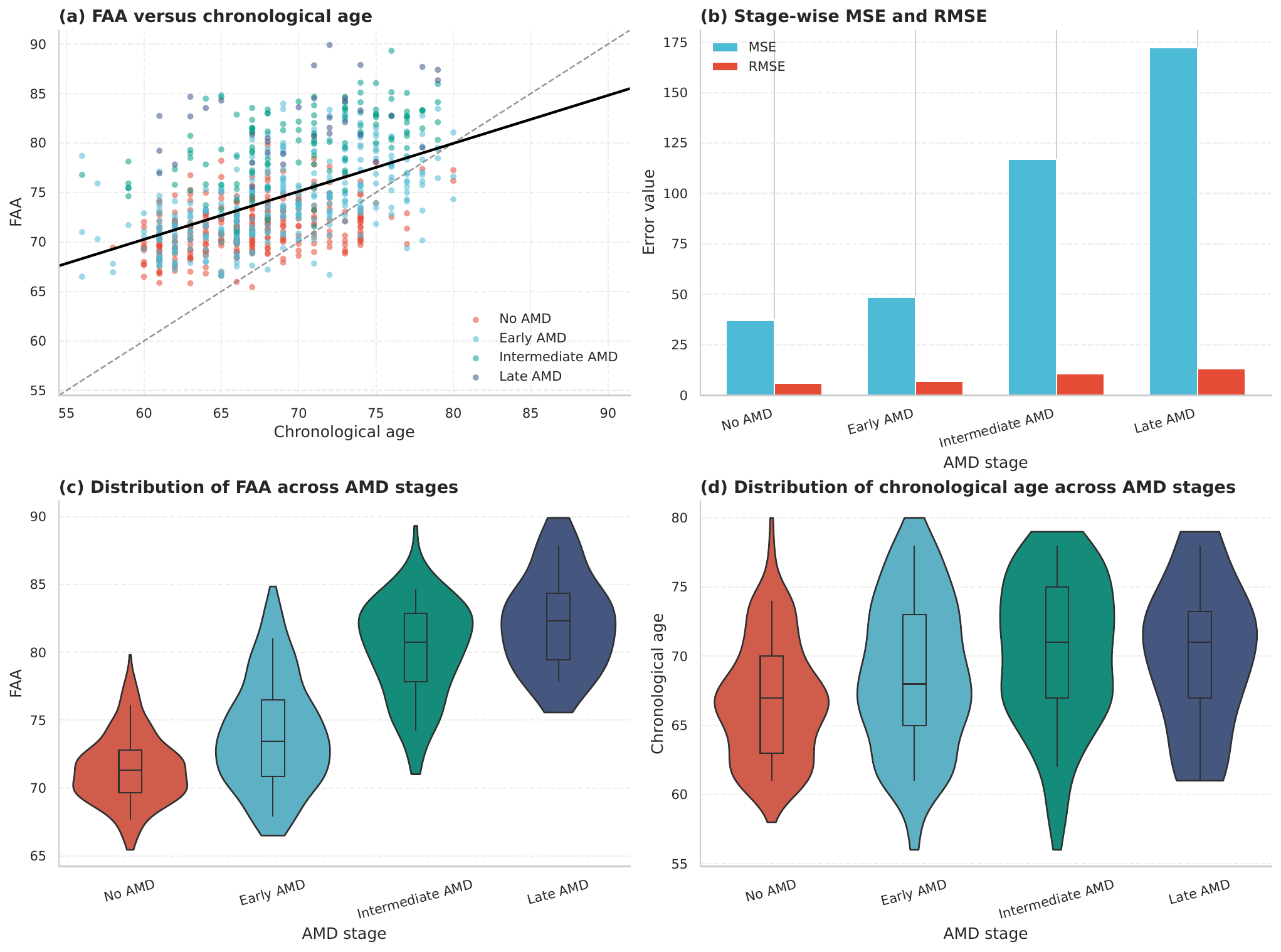}
\caption{\textbf{Characterization of FAA in relation to chronological age and AMD severity.} \textbf{a}, Relationship between chronological age and fundus-based AMD age (FAA) in the testing cohort. Each point represents one eligible eye and is colored according to baseline AMD stage. The dashed diagonal line represents perfect agreement between FAA and chronological age, and the solid line shows the fitted linear trend. \textbf{b}, Mean squared error (MSE) and root mean squared error (RMSE) of FAA relative to chronological age across baseline AMD stages. \textbf{c}, Distribution of FAA across no AMD, early AMD, intermediate AMD, and late AMD stages. \textbf{d}, Distribution of chronological age across no AMD, early AMD, intermediate AMD, and late AMD stages. The stage-wise distribution of FAA is consistent with the severity-guided ranking objective.}
\label{fig:faa relationship}
\end{figure}

\zcref{fig:faa relationship} characterizes the relationship of FAA with chronological age and baseline AMD severity in the held-out AREDS internal-test cohort. FAA was positively associated with chronological age (\zcref{fig:faa relationship}a), indicating that the learned representation retained chronological-age-related retinal information while not being identical to chronological age itself. The error of FAA relative to chronological age varied across AMD stages (\zcref{fig:faa relationship}b). Both the mean squared error and root mean squared error increased from the no-AMD and early-AMD groups to the intermediate-AMD and late-AMD groups, indicating larger deviations from chronological age in eyes with more advanced AMD-related retinal changes. FAA also showed a stage-wise increase across the no-AMD, early-AMD, intermediate-AMD, and late-AMD groups (\zcref{fig:faa relationship}c), whereas chronological age alone showed a more modest distributional shift across these stages (\zcref{fig:faa relationship}d). Compared with chronological age, FAA showed greater differentiation across AMD stages, supporting its stronger association with AMD-related retinal changes.

\begin{figure}[t!]
\centering
\includegraphics[width=\linewidth]{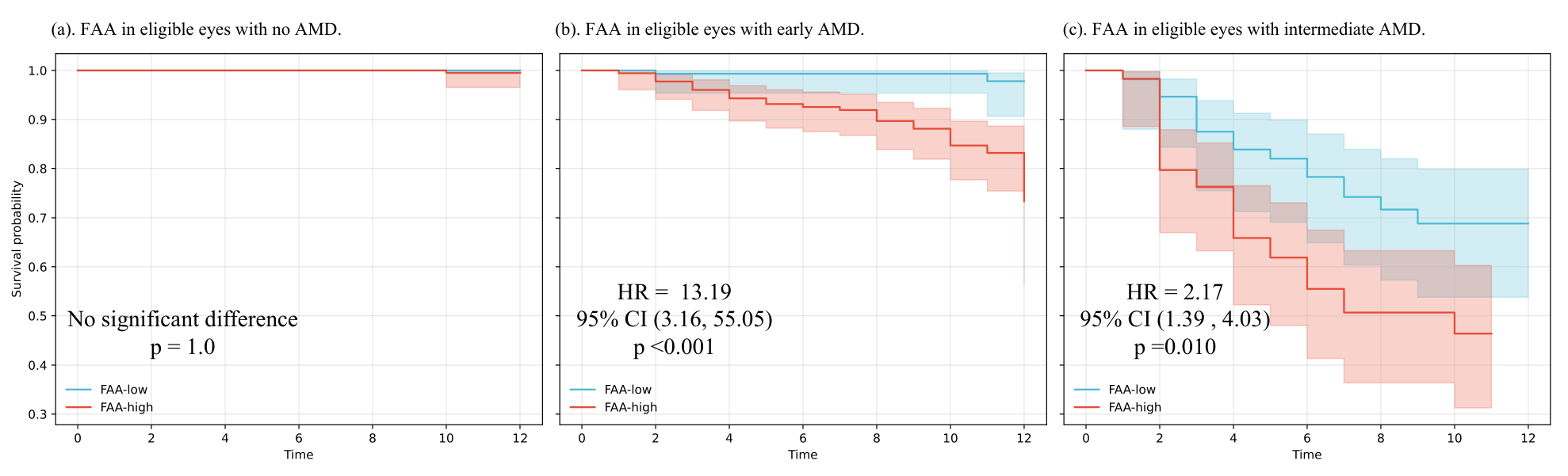}
\caption{\textbf{Kaplan--Meier analysis of AMD progression stratified by FAA.} Eligible eyes were categorized into high- and low-risk groups using the median FAA within each AMD stage.
\textbf{a}, No AMD: no significant difference ($p = 1.0$). \textbf{b}, Early AMD: higher FAA is associated with increased late AMD risk (HR = 13.19, 95\% CI 3.16--55.05, $p<0.001$). \textbf{c}, Intermediate AMD: similar separation is observed (HR = 2.17, 95\% CI 1.39--4.03, $p = 0.010$). P values displayed in the Kaplan--Meier panels were obtained separately using two-sided log-rank tests.
}
\label{fig:km strratified}
\end{figure}

To evaluate the prognostic value of FAA, we conducted Kaplan--Meier analyses across no, early, and intermediate AMD subgroups, using FAA as the sole stratification variable. For each stage, higher- and lower-FAA groups were defined using the median FAA among development eyes within that stage, and the resulting cutoff was applied unchanged to internal-test eyes in the corresponding stage (\zcref{fig:km strratified}). Higher FAA was consistently associated with increased risk of late AMD progression, with the strongest effect observed in early and intermediate stages. In the early AMD stage, the high-FAA group showed significantly reduced progression-free survival (HR = 13.19, 95\% CI 3.16--55.05, $p<0.001$), and a similar pattern was observed in the intermediate AMD stage (HR = 2.17, 95\% CI 1.39--4.03, $p = 0.010$). No significant separation was observed in the no-AMD stage ($p = 1.0$), likely due to low event rates in this subgroup. Collectively, these results demonstrate that FAA alone provides effective risk stratification among the eyes identified as having early or intermediate AMD.

\subsection{Interocular correlation and eye-specific variation in FAA}\label{interocular-correlation-and-eye-specific-variation-in-faa}

\begin{figure}[t!]
\centering
\includegraphics[width=1.0\linewidth]{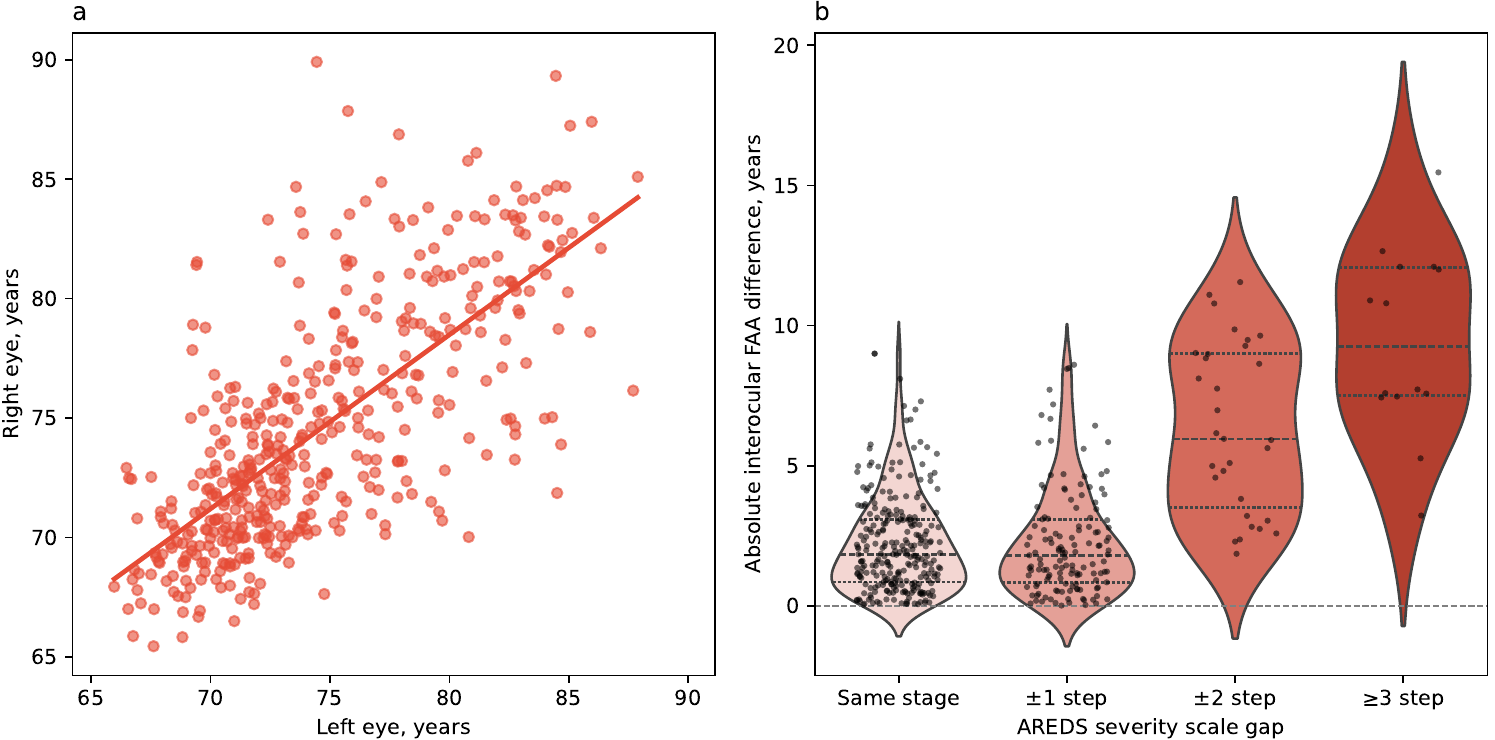}
\caption{\textbf{Interocular correlation and eye-specific variation in FAA.} \textbf{a}, Correlation between FAA estimates from the left and right eyes of the same participant. \textbf{b}, Distribution of the absolute interocular FAA difference across categories of the absolute difference in baseline AMD stages between fellow eyes.
}
\label{fig:interocular correlation}
\end{figure}

\zcref{fig:interocular correlation}a shows the correlation of FAA between the left and right eyes of the same participant. The observed correlation indicates substantial interocular concordance and within-participant dependence (Pearson $r \approx 0.72$). This finding indicates substantial interocular correlation and highlights the within-participant dependence of fellow-eye observations.

We next examined eye-specific variation in FAA in relation to baseline interocular AMD severity differences. \zcref{fig:interocular correlation}b shows the distribution of the absolute interocular FAA difference (defined as the absolute difference between the left- and right-eye FAA estimates) grouped by AMD stage. Eye pairs with the same or similar stages generally showed smaller FAA differences, whereas greater stage differences between fellow eyes were accompanied by larger interocular FAA differences.

\subsection{Illustrative Case Analysis}\label{illustrative-case-analysis}

\begin{figure}[t!]
\centering
\includegraphics[width=\linewidth]{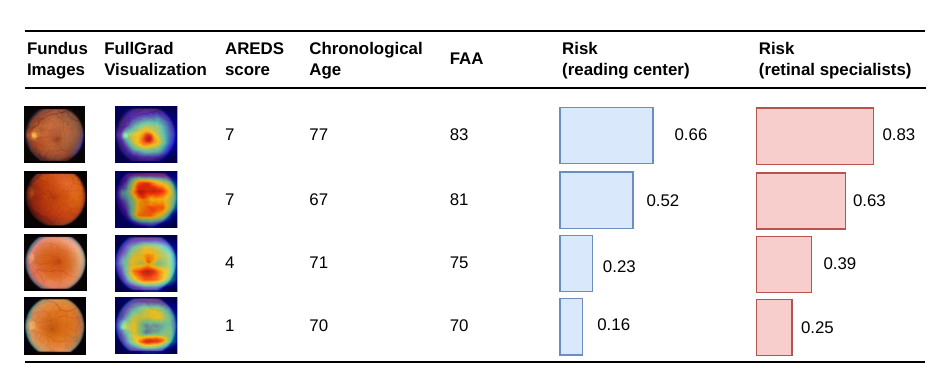}
\caption{\textbf{Illustrative examples of AMD risk predictions.} Each case includes a color fundus photograph, FullGrad Visualization~\cite{srinivas2019full}, AMD severity score, chronological age, FAA, and the associated predicted AMD risk score from both the reading-center-defined and the specialist-defined Cox models. Survival outcomes were as follows: \textbf{a}, developed late AMD after 3 years, \textbf{b}, developed late AMD after 7 years, \textbf{c}, developed late AMD after 12 years, and \textbf{d}, remained undiagnosed with late AMD after 12 years. Risk values were derived from the reading-center-defined/ specialist-defined Cox model and normalized to the range [0, 1] for visualization.}
\label{fig:examples}
\end{figure}

To further demonstrate the practical value and clinical interpretability of our AMD risk prediction framework, we provide illustrative examples that include color fundus photographs, corresponding AMD severity scores, chronological ages, FAA, and normalized model-specific risk scores derived from both reading-center-defined and specialist-defined Cox models. We additionally incorporate FullGrad visualizations~\cite{srinivas2019full}, which highlight retinal regions that contribute to the FAA estimation. These maps reveal that the FAA focuses on physiologically meaningful structures, such as the macular region, drusen-related abnormalities, and pigmentary changes, confirming that the highlighted regions are consistent with regions known to be affected in AMD, particularly the macular area, and supporting the model's interpretability. As shown in \zcref{fig:examples}, FAA aligns closely with the AMD severity score and progression timelines, capturing disease-relevant retinal patterns associated with AMD risk that may not be apparent when considering chronological age alone. Differences between the reading-center-defined and specialist-defined risk predictions highlight the influence of phenotype definitions on risk estimation, reflecting varying levels of detail and specificity in phenotypic assessments. Overall, these examples highlight the enhanced prognostic capabilities provided by integrating FAA and feature-level interpretability via FullGrad into AMD risk prediction frameworks.

\subsection{Comparison with existing benchmarks and traditional clinical predictors}\label{comparison-with-existing-benchmarks-and-traditional-clinical-predictors}

To further evaluate the performance and independence of the proposed \textbf{FAA}, we conducted three complementary analyses.

\begin{figure}[t!]
\centering
\includegraphics[width=\linewidth]{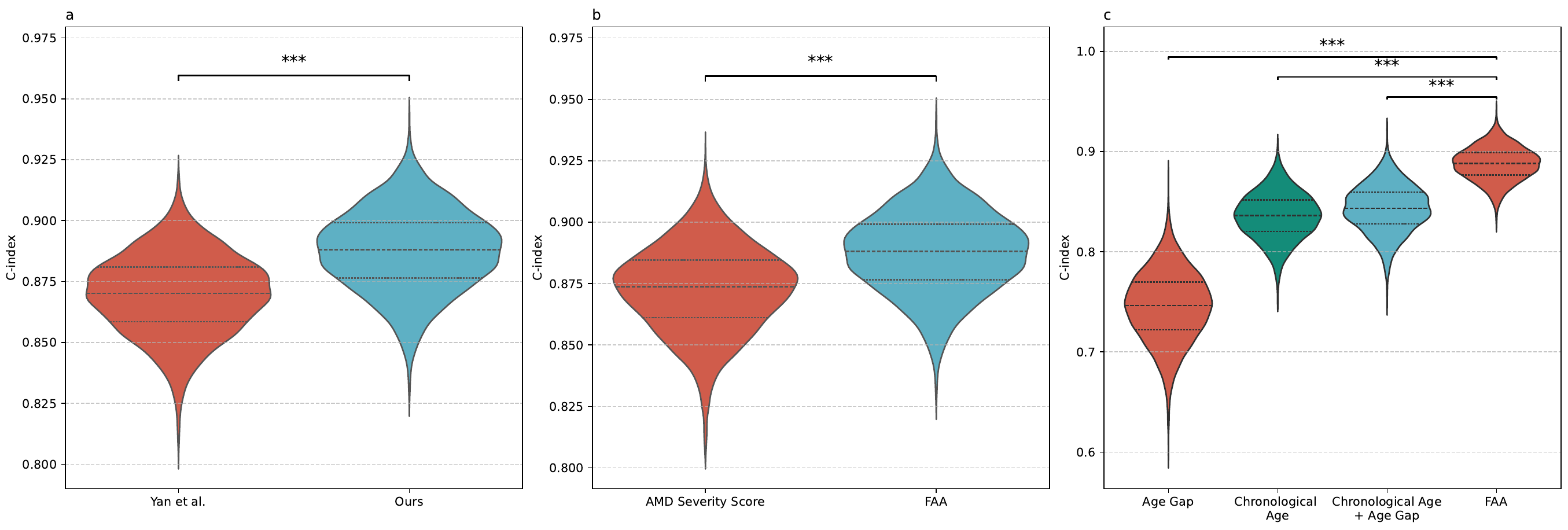}
\caption{\textbf{Comparative analysis of the proposed FAA against existing benchmarks}. \textbf{a}, An adapted image-only benchmark based on Yan et al. \textbf{b}, AMD severity grading. \textbf{c}, Using age gap, chronological age, chronological age + age gap, and FAA as inputs. P-values were calculated using two-sided bootstrap resampling (5,000 iterations). Statistical significance was indicated as *** for $p<0.001$.}
\label{fig:against benchmarks}
\end{figure}

We first compared our model against a representative deep learning--based AMD progression prediction model proposed by~\citet{yan2020deep} (\zcref{fig:against benchmarks}a). The benchmark was implemented and evaluated under the matched AREDS data partition and 5-year progression setting described under \zcref{ssec:benchmark}. The FAA-based model achieved a higher C-index than the adapted benchmark (0.888 [95\% CI, 0.853--0.919] vs. 0.870 [95\% CI, 0.836--0.899]). These results suggest that modeling a disease-specific retinal age representation may provide more discriminative prognostic information than direct image-to-risk frameworks under the matched evaluation setting.

We compared the FAA with a Cox regression model that used only the AMD Severity Score as input. As shown in \zcref{fig:against benchmarks}b, the proposed model achieved a higher C-index than the severity-score-only model (C-index 0.888 [95\% CI: 0.853-0.919] vs. 0.873 [95\% CI: 0.836-0.905]). These results confirm that FAA captures prognostic information beyond conventional AMD severity grading and generalizes beyond handcrafted clinical labels.

Here, we define the fundus-based AMD age gap as the difference between FAA and chronological age. As shown in \zcref{fig:against benchmarks}c, we compared models that incorporated age gap, chronological age, chronological age + age gap, or FAA as the age-related input variable in the Cox model. The FAA-based model achieved the highest C-index (0.888 [95\% CI: 0.853-0.919]) and significantly outperformed other models including age-gap based model (0.745 [95\% CI: 0.674--0.810]), chronological-age based model (0.836 [95\% CI: 0.788-0.880]) and chronological-age based model incorporating age gap (0.843 [95\% CI: 0.798-0.888]), confirming that FAA captures disease-specific retinal aging signals that cannot be explained by general aging indicators.

\subsection{Ablation Study: Sensitivity Analysis of Model Components}\label{ablation-study-sensitivity-analysis-of-model-components}

To evaluate the robustness and contribution of individual model components, we conducted ablation studies on both the loss function design and the backbone architecture.

\begin{figure}[t!]
\centering
\includegraphics[width=\linewidth]{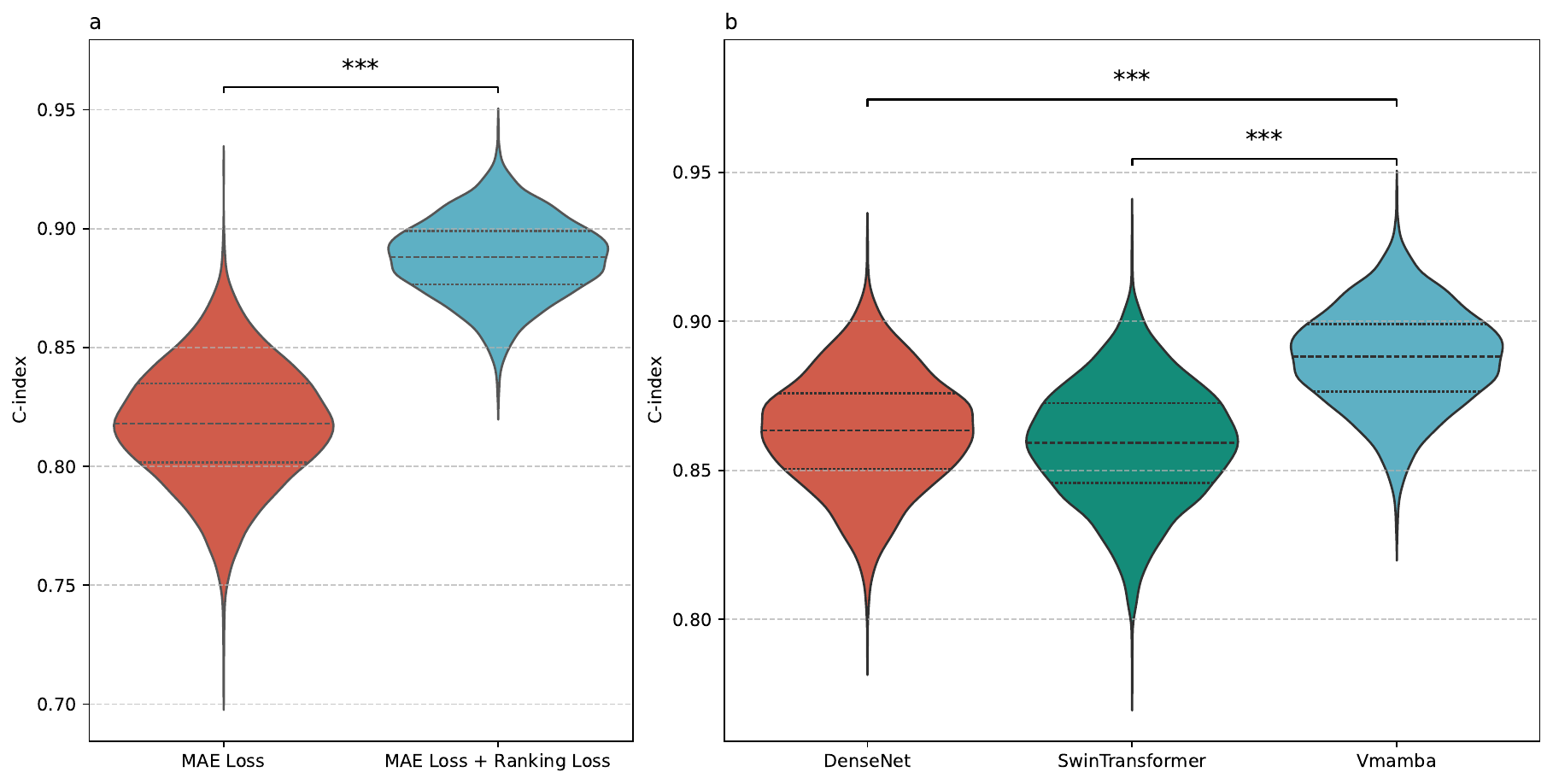}
\caption{\textbf{Ablation study on different losses and backbones. We used four phenotypes, as defined by the reading center, to assess the prognostic performance using the C-index.} \textbf{a}, Comparison of different loss functions. \textbf{b}, Comparison of different backbones. P-values were calculated using two-sided bootstrap resampling (5,000 iterations). Statistical significance was indicated as *** for $p<0.001$.}
\label{fig:ablation}
\end{figure}

We compared our proposed ranking loss, combined with the mean absolute error (MAE) loss, against the MAE loss alone. \zcref{fig:ablation}a indicates that incorporating the ranking loss substantially enhanced model performance. Specifically, the reading-center-defined Cox model has a significant performance improvement with the proposed ranking loss (0.888 [95\% CI:0.853-0.919]) compared to using the MAE loss only (0.818 [95\% CI:0.767-0.865], $p<0.001$). These findings demonstrate the effectiveness of our ranking loss in guiding the model to better generalize across varying stages of AMD.

We assigned different backbone architectures, including a traditional convolutional neural network\\
(DenseNet)~\cite{huang2017densely}, a Transformer-based model (SwinTransformer)~\cite{liu2021swin}, and the VMamba architecture. \zcref{fig:ablation}b shows that the VMamba backbone outperformed both DenseNet (0.864 [95\% CI: 0.825-0.899], $p<0.001$) and SwinTransformer (0.859 [95\% CI: 0.818-0.896], $p<0.001$), demonstrating a significantly higher C-index than CNN and Transformer. These results highlight the effectiveness of the VMamba architecture in capturing both local and global retinal features in FAA estimation and AMD risk prediction.

\section{Discussion}\label{discussion}

Our study introduces an approach to predict AMD progression risk by integrating fundus-based AMD age (FAA) with established AMD phenotypes. We found that our late AMD progression prediction model, which incorporates FAA into Cox proportional hazards models, significantly improved the predictive accuracy for AMD progression compared to models using chronological age alone. FAA outperforms chronological age because it integrates fundus-derived retinal features and detailed phenotypic annotations, thereby providing a more comprehensive and precise indicator of retinal health status. Unlike chronological age, which is a general measure of aging, FAA effectively captures individualized differences in retinal pathology and functional deterioration associated with AMD. Notably, this study validated the effectiveness of the proposed FAA-based prediction model across two large, multicenter datasets (AREDS and AREDS2), demonstrating its generalizability and clinical applicability. Our findings emphasize that the FAA provides a more accurate reflection of the underlying retinal aging processes relevant to AMD, thus enhancing risk assessment and clinical decision-making. Moreover, the consistent performance improvements observed across different phenotype definitions further underscore the applicability and reliability of FAA as a prognostic biomarker for AMD.

From a clinical perspective, FAA offers a practical and interpretable biomarker that could assist ophthalmologists in risk stratification, follow-up scheduling, and early intervention planning. By providing a reasonable ``retinal age'', the FAA enables clearer communication of disease risk to patients and supports more personalized management strategies. Clinicians can use FAA to more accurately estimate each patient's likelihood of progressing to late AMD, thereby identifying individuals who may require closer monitoring or proactive treatment. As illustrated in \zcref{fig:km others}, incorporating FAA with other clinical risk factors (e.g., chronological age, smoking status, and AMD phenotypes) within a Cox proportional hazards framework allows the model to more effectively separate patients into \textbf{high-risk and low-risk subgroups}. Patients classified as \textbf{high-risk} demonstrate markedly lower survival probabilities, suggesting that this framework could help ophthalmologists prioritize follow-up schedules and implement targeted early interventions for those most likely to progress to disease progression.

Although the concept of biological aging has long been acknowledged, there has been a recent surge in public and scientific interest in its clinical implications. It is now widely recognized that healthcare decisions should not be based solely on chronological age; instead, there is increasing consensus that they should also consider the cumulative physiological and psychological effects. While cellular and molecular ``frailomics'' measures have been widely explored as indicators of biological aging, imaging biomarkers have received comparatively little attention for their potential in this domain. In this study, we hypothesized that features derived from color fundus photographs could provide a macroscopic representation of age-related and AMD-related retinal variation. Across analyses using phenotypes graded by the Reading Center or assessed by retinal specialists, Cox models incorporating FAA showed better prognostic performance than the corresponding models using chronological age. These findings support further investigation of disease-specific retinal age representations for AMD risk stratification. Prospective validation and clinically defined decision thresholds will be needed before such measures can be used to identify intervention windows or guide preventive management.

Previous biological age estimation studies have typically been trained on healthy retinal images~\cite{yu2023retinal,liu2019biological,zhu2023retinal,park2024artificial}. Thus, they tended to perform poorly when applied to images with AMD, due to substantial anatomical and pathological differences. To overcome this limitation, we proposed a ranking loss guided by an effective 10-level ranking label for FAA training. As shown in \zcref{fig:ablation}, this strategy significantly enhanced the model's ability to recognize and adapt to the unique visual characteristics of various AMD stages, thereby enabling more accurate retinal age estimation across the spectrum of AMD progression.

Previous studies have typically employed deep learning--based AMD prognosis models that directly predict disease risk from fundus images in a black-box manner~\cite{holste2024harnessing,yan2020deep,peng2020predicting,gervelmeyer2024interpretable}. To further evaluate our contribution, we compared our FAA-based model with a benchmark model proposed by~\citet{yan2020deep}. As shown in \zcref{fig:against benchmarks}a, our model achieved a notably higher C-Index than the adapted benchmark. These results suggest that modeling disease-specific retinal aging provides more discriminative prognostic information than image-to-risk frameworks.

Previous studies have frequently evaluated the difference between image-predicted age and chronological age as a standalone biomarker. In this study, FAA age gap (defined as FAA minus chronological age) had distinct roles in model development and secondary analyses. During FAA development, it was calculated from the network output and chronological age and used in the severity-guided ranking objective. The age gap itself does not contain intrinsic age information and is, therefore, insufficient as a direct replacement for FAA. For instance, as shown in \zcref{fig:examples}, individuals aged $\ge 70$ years may exhibit small age gaps but still carry an elevated AMD risk due to their inherently higher chronological age. Our ablation studies (\zcref{fig:against benchmarks}c) further substantiate this point, demonstrating that models using FAA alone achieve the best predictive performance. Although incorporating both chronological age and the age gap yields modest improvements, this approach does not surpass the model that uses FAA alone, underscoring FAA's superior predictive value for assessing AMD progression risk.

Moreover, our study leveraged two sets of phenotypes--specialist-defined and reading-center-defined. Retinal specialist-defined phenotypes derived from routine diagnostic procedures without specialized equipment tend to be less precise than those provided by human graders at the reading center. Consequently, our models demonstrated slightly lower prognostic performance when using specialist-defined phenotypes than when using reading-center-defined phenotypes. Nonetheless, the model maintained strong predictive capabilities in both scenarios, affirming the robustness and practical applicability of our proposed methodology regardless of the source of phenotype definition.

Because FAA is guided by AMD severity during training, a potential concern is that it may simply re-encode existing AMD severity scores. To address this directly, we compared the FAA-based model with a Cox model using only the AMD Severity Score as input. As shown in \zcref{fig:against benchmarks}b, our FAA-based model achieved a higher C-index than this severity-score-only baseline, indicating that it captures prognostic information beyond conventional severity grading. FAA showed a stage-wise increase across baseline AMD severity groups, which is expected in part because AMD severity contributes directly to the ranking objective used during FAA training. We therefore interpret \zcref{fig:faa relationship} as a descriptive characterization of the learned FAA representation rather than as independent prognostic validation. Evidence that FAA provides prognostic information beyond chronological age comes instead from the held-out model comparisons, nested Cox-model analyses, and the observed-progression analyses presented in \zcref{fig:prognostic performance, fig:prognostic performance human, fig:km strratified}. Specifically, Kaplan--Meier analyses (\zcref{fig:km strratified}) show that FAA alone separates individuals into distinct risk groups when identified as having early or intermediate AMD. This within-stage separation suggests that FAA captures disease heterogeneity that is not fully reflected by conventional grading systems.

This study has several limitations. First, differences in phenotypic definitions between the AREDS and AREDS2 datasets may affect the direct comparability of risk prediction models. To mitigate this, we harmonized phenotypes in AREDS2 to match the definitions used in AREDS (\zcref{ssec:harmonization}). However, this approach requires access to compatible phenotype annotations, which may not always be available in other datasets. The FAA estimator, which uses CFPs alone at inference, was applied to AREDS2 without retraining. In this related external clinical-trial cohort, the FAA-based models retained higher discrimination than the corresponding chronological-age-based models, supporting further cross-cohort evaluation of the FAA representation. In addition, FAA values used for Cox-model development were obtained by averaging predictions from all five models, which may have introduced optimistic bias during Cox-model fitting. Nevertheless, no participant from the AREDS internal-test or AREDS2 cohort was included in FAA training or validation, and all reported held-out performance metrics were evaluated in participants whose complete image histories were unseen by the FAA models. Future studies should use participant-level cross-fitting or nested model-development procedures when generating FAA values for downstream survival-model fitting.

Second, our study did not separately analyze the two distinct subtypes of late AMD, namely geographic atrophy and neovascular AMD. This decision reflects our primary aim of stratifying risk and guiding early interventions for individuals with no, early, or intermediate AMD. Therefore, individuals who had already progressed to late-stage AMD at the baseline visit were outside the scope of our work. Future studies may benefit from exploring subtype-specific prognostic markers to further refine risk prediction and tailor management strategies.

Third, although we showed interpretability by adding FullGrad visualizations in \zcref{fig:examples}, interpretability remains a limitation of our model. The updated examples show that the FAA estimation network highlights retinal regions that correspond well with clinically relevant AMD patterns, supporting the model's predictions. However, we acknowledge that FullGrad visualizations alone do not fully resolve the ``black-box'' nature of deep state-space models. The highlighted regions primarily localize to the macular area and should be interpreted as coarse localization rather than precise identification of fine-grained pathological features. The mechanisms underlying the identification of retinal characteristics associated with aging and AMD risk remain incompletely understood. Future work will incorporate more advanced, dedicated explainability tools tailored to state-space architectures to more comprehensively elucidate the model's decision-making process and further enhance clinical interpretability.

Fourth, both AREDS and AREDS2 were U.S.-based clinical-trial cohorts composed predominantly of White participants. Consequently, external validation in AREDS2 does not establish transportability across racial and ethnic groups, geographic regions, or routine-care settings. Population differences in retinal characteristics, AMD risk profiles, and healthcare contexts may affect model performance. Future studies should evaluate FAA in racially, ethnically, and geographically diverse cohorts and determine whether population-specific recalibration is required before broader clinical implementation.

In addition, FAA has not yet been validated against external biological aging benchmarks, such as molecular aging markers, systemic frailty indices, or multimodal imaging biomarkers. Because such measurements are not available in AREDS/AREDS2, our analysis is limited to retinal aging signals. Future studies incorporating molecular or multimodal aging assessments will be important for further validating the biological relevance and interpretability of FAA.

In summary, we have demonstrated that incorporating retinal-based FAA into AMD risk prediction models significantly enhances predictive accuracy and clinical relevance compared to traditional models that rely solely on chronological age. Despite inherent limitations, such as differences in phenotypic definitions across datasets and the lack of subtype-specific analysis for late-stage AMD, our approach showed reproducible prognostic improvements across the two study cohorts and warrants further evaluation for potential clinical application. Future research addressing these limitations and validating the FAA in diverse populations could further solidify its role as a pivotal biomarker in the management and prognosis of AMD.

\section{Methods}\label{methods}

\subsection{Dataset Description}
\label{sec:dataset-description}

The present study used data from the Age-Related Eye Disease Study (AREDS and AREDS2)~\cite{study1999age,chew2012age}. Both are large-scale, longitudinal clinical trial datasets sponsored by the National Eye Institute (NEI) and designed primarily to investigate the clinical course, prognosis, and potential treatments for AMD (\zcref{tab:cohort-characteristics}). The AREDS and AREDS2 datasets include detailed demographic information, lifestyle factors, extensive ophthalmological examinations, retinal imaging, and genetic data collected systematically over long-term follow-up. Due to their comprehensive nature and standardized data-collection protocols, AREDS and AREDS2 have become a foundational resource for numerous studies aimed at understanding AMD progression, risk factors, and evaluating predictive biomarkers. Throughout the manuscript, ``participant'' refers to a unique study participant; ``longitudinal CFP image'' refers to an eye- and visit-specific image acquisition; ``baseline eye record'' refers to an eye represented at the qualifying baseline visit before endpoint- and phenotype-specific eligibility criteria were applied; and ``analytical eye'' refers to a final eligible eye included in a specific survival-model analysis.

The AREDS cohort comprised 2,740 participants and was initially divided into a development cohort of 2,192 participants and an internal test cohort of 548 participants. The 2,192 AREDS development participants were partitioned into five mutually exclusive folds at the participant level. Five FAA models were trained, each using four folds for training and the remaining fold for validation. All images from both eyes and all longitudinal visits of a participant were retained within the same fold. Subsequently, the predicted FundusAge was combined with AMD phenotypes to develop a prognostic Cox proportional hazards model~\cite{cox1972regression}for predicting AMD progression. Of the 548 participants assigned to the internal-test cohort, 449 contributed baseline eye records to the baseline analytical dataset, yielding 898 baseline eye records before endpoint- and phenotype-specific eligibility criteria were applied. After applying the eligibility criteria, the final Cox analytical samples included 3,799 development eyes and 769 internal-test eyes for the reading-center-defined endpoints, and 3,840 development eyes and 811 internal-test eyes for the specialist-defined endpoints. All dataset partitioning was performed at the participant level using unique participant identifiers. Images from both eyes, repeated acquisitions, and all longitudinal visits of the same participant were retained within the same partition.

Additionally, as shown in \zcref{fig:overview}b, we incorporated data from the independent AREDS2 dataset, consisting of 4,192 participants and 26,215 retinal images. AREDS2 was used exclusively as an independent external-validation cohort. No AREDS2 data were used for FAA training, Cox-model fitting, model selection, hyperparameter optimization, or risk-threshold determination. Among the baseline images, 6,754 eyes had valid FAA estimates, of which 6,548 were free of late AMD at baseline. After applying phenotype-specific completeness criteria and prespecified participant-level exclusions, the final external-validation samples comprised 5,651 eyes for the reading-center-defined analysis and 5,964 eyes for the specialist-defined analysis.

\begin{table}[t!]
\centering
\caption{\textbf{Summary of demographic and clinical characteristics of training and testing datasets used in this study.} The AREDS internal-test partition comprised 548 participants and 9,036 longitudinal CFP images; 449 of these participants contributed baseline eye records and were included in the baseline demographic summary. Data are presented as mean $\pm$ standard deviation or counts, as appropriate. {GEOAWI}: Geographic atrophy within the grid. {RPEDWI}: RPE depigmentation area within the grid. {INCPWI}: Increased pigment area within the grid. {DRARWI:} Drusen area within the grid. {DRUSENQ}: clinical drusen size assessment. {PIGDRUQ}: pigment abnormalities. Detailed definitions and scoring criteria for each phenotype are provided in \zcref{ssec:amd definition}.}
\label{tab:cohort-characteristics}
\resizebox{\textwidth}{!}{%
\begin{tabular}{llll}
\toprule
& \multicolumn{2}{l}{AREDS}
& AREDS2 \\
\cmidrule(lr){2-3}
Characteristic & Training Set & Testing Set & Testing Set \\
\midrule
\rowcolor{gray!20}\multicolumn{4}{l}{\textit{Participant characteristics}} \\
Number of participants
& 2,192 & 449 & 4,192 \\
Baseline age (mean $\pm$ SD, years)
& $68.74 \pm 5.12$ & $68.28 \pm 5.10$ & $73.13 \pm 7.75$ \\
Smoking history (never/former/current)
& 1,063/1,000/129
& 184/231/34
& 1,820/2,091/281 \\
Race (White/Black/Asian/Others)
& 2,112/66/4/10
& 419/26/1/3
& 4,055/50/30/57 \\
Follow-up, median (IQR), years
& 10.0 (6.0--11.0)
& 10.0 (7.0--11.0)
& 5.0 (3.0--5.0) \\
\rowcolor{gray!20}\multicolumn{4}{l}{\textit{Image characteristics}} \\
Number of longitudinal CFP images
& 36,782 & 9,036 & 26,215 \\
AMD severity scale score (no/early/intermediate/late)
& 14,212/11,607/6,732/4,231
& 3,496/2,891/1,486/1,163
& 92/2,844/15,130/8,149 \\
AMD severity scale score at baseline visit
& 1,766/1,389/821/407
& 350/347/131/70
& 49/1,299/5,200/206 \\
\rowcolor{gray!20}\multicolumn{4}{l}{\textit{Specialist-defined phenotypes}} \\
DRUSENQ (0/1/2/3)
& 557/1,340/624/1,319
& 235/255/123/198
& 31/150/5,779/4 \\
PIGDRUQ (0/1)
& 3,036/804
& 582/229
& 2,905/3,059 \\
\rowcolor{gray!20}\multicolumn{4}{l}{\textit{Expert-defined phenotypes}} \\
INCPWI (0--7)
& 2,853/98/82/198/307/208/37/16
& 591/22/12/52/59/31/0/2
& 1,839/182/2,132/1,498 \\
RPEDWI (0--7)
& 3,277/185/103/106/45/37/26/20
& 665/34/24/21/9/6/6/4
& 3,971/199/525/697/259 \\
GEOAWI (0--7)
& 3,745/7/2/2/7/13/8/15
& 762/3/0/2/1/1/0/0
& 5,402/0/4/55/70/43/37/40 \\
DRARWI (0--7)
& 973/851/603/335/308/224/176/329
& 157/126/121/83/99/75/39/69
& 46/32/136/202/569/999/1,389/2,278 \\
\rowcolor{gray!20}\multicolumn{4}{l}{\textit{Progression to late AMD (classified by Reading Center)}} \\
Late AMD: \% of eyes at year 0/1/2/3/4/5/all years
& 
9.3/10.0/11.8/12.8/13.9/14.9/19.8
&
7.8/8.1/10.3/11.2/12.8/13.3/16.6
&
2.4/9.0/14.4/20.4/26.0/29.9/30.5 \\
\bottomrule
\end{tabular}%
}
\end{table}

\subsection{Phenotype, outcome, and censoring definitions}\label{phenotype-outcome-and-censoring-definitions}

Two separate Cox proportional hazards models were built using different sets of AMD phenotypes in the AREDS dataset. One set of phenotypes was defined by expert image graders in the reading center, including geographic atrophy (GEOAWI), retinal pigment epithelium depigmentation (RPEDWI), increased pigment (INCPWI), and drusen area (DRARWI), each scored within the central grid on a 0-7 scale. The other set was defined by retinal specialists in clinical practice and consisted of clinical drusen size assessment (DRUSENQ, 0-3) and pigment abnormalities (PIGDRUQ, yes/no). This design allowed us to validate the predictive value of FAA under distinct phenotype annotation standards.

Phenotype harmonization was performed specifically for the AREDS2 external-validation analysis. The primary AREDS internal-test Cox models were fitted and evaluated using the native AREDS phenotype coding. For external validation, corresponding variables were mapped to common, variable-specific numerical scales. AREDS RPEDWI and INCPWI were coarsened from their native 0--7 scales to 0--4 and 0--3, respectively, to match the grading resolution of AREDS2 DPINGRID and IPINGRID. Conversely, the continuous AREDS2 GADAGRID variable was discretized according to the native 0--7 AREDS GEOAWI thresholds. Variables that already shared compatible coding were retained without transformation. The corresponding harmonized Cox models were fitted using only the AREDS development cohort and subsequently applied to AREDS2 without model fitting, recalibration, or outcome-based tuning in AREDS2. The complete variable mappings are provided in \zcref{tab:supp-cox-variables}.

All covariates included in the Cox proportional hazards models were defined at baseline. Continuous variables, including fundus-based AMD age (FAA) and chronological age, were modeled per 1-year increase. AMD phenotype variables (e.g., GEOAWI, RPEDWI, and INCPWI) were treated as ordinal variables and entered the Cox model as continuous scores based on their grading scales. Hazard ratios for these variables, therefore, correspond to a one-grade increase in the respective phenotype score. Categorical variables, including smoking status, were encoded using indicator (dummy) variables, with ``never smoker'' as the reference category. No continuous interpretation or unit was assigned to categorical variables. Detailed definitions and coding schemes of all variables are provided in \zcref{tab:supp-cox-variables}.

The AREDS data used in this study contained an eye-specific 12-level severity coding derived from the AREDS detailed severity system. Levels 1--9 represented ordered non-late AMD severity, whereas levels 10--12 represented late-AMD categories. For descriptive reporting and stage-stratified analyses, we grouped these labels as no AMD (score 1), early AMD (scores 2--5), intermediate AMD (scores 6--9), and late AMD (scores 10--12). This four-category grouping was not used to train the FAA estimation model or fit the primary Cox models.

Progression to late AMD was defined at the eye level. Eyes with an original AREDS severity score of 10--12 at baseline were excluded from survival modeling because they had already reached the study endpoint. For each eligible eye with a baseline severity score of 1--9, follow-up time was measured from the baseline visit to the first follow-up visit at which that eye reached a late-AMD severity category, defined as an original AREDS severity score of 10--12. Eyes that developed late AMD during follow-up were coded as events. Eyes without observed progression were right-censored at their last available gradable ophthalmic visit. Thus, the follow-up time represented either the time from baseline to the first observed late-AMD event or, for a censored eye, the time from baseline to its last available gradable follow-up examination.

Mortality was not available as a separately adjudicated competing event in the analytical dataset, and the reason for the termination of follow-up was not separately modeled. Death was therefore not treated as a competing event using a cause-specific or distribution hazard model. Eyes without an observed late-AMD event were right-censored at their last available gradable ophthalmic visit, and the survival analyses relied on the standard independent-censoring assumption. Consequently, potentially informative censoring associated with death, withdrawal, or loss to follow-up could not be excluded.

Missing data were handled using complete-case analysis separately within each phenotype-definition setting. Eyes missing follow-up time, event status, FAA, chronological age, smoking status, or any phenotype required for the corresponding Cox model were excluded. No statistical imputation was performed. Within each phenotype-definition setting, the FAA-based model, chronological-age-based model, and corresponding comparison models were fitted and evaluated using the same complete-case analytical sample.

\subsection{Pipeline of proposed method}
\label{pipeline-of-proposed-method}

We developed a comprehensive pipeline to predict AMD progression risk by integrating deep learning-derived FAA into a survival analysis framework (\zcref{fig:overview}c). Traditionally, a Cox proportional hazards model is constructed using chronological age, smoking status, and relevant AMD phenotypes defined by specialists or experts to predict AMD progression risk. In contrast, our proposed method replaces chronological age with a deep-learning-estimated FAA derived directly from CFPs. Specifically, we first trained the FAA estimation model using the entire set of available images from the training cohort (2,192 participants, 36,782 images). Subsequently, for training and testing the Cox model, we used only images from each participant's baseline visit. We also excluded cases already identified with late AMD (AMD severity score \(\geq 10\)), as such cases are clinically considered to have already reached the progression endpoint. The proposed pipeline ensures accurate evaluation of the FAA's role in predicting AMD progression among patients with normal, early, and intermediate stages of AMD, enabling better patient risk stratification and clinical management.

\subsection{FAA model development}\label{faa-model-development}

Fundus-based AMD age (FAA) was defined as an AMD-related, fundus-based age derived from a color fundus photograph (CFP). For image \(x_{i}\), the FAA model produced one continuous estimate \({\hat{a}}_{i} = f_{\theta}\left( x_{i} \right)\), where \(a_{i}\) denotes the corresponding chronological age. Participant identifiers, visit numbers, temporal ordering, longitudinal image sequences, survival times, and future progression outcomes were not provided as model inputs. All participant-level dataset allocation and five-fold partitioning were performed using unique participant identifiers, with images from both eyes and all longitudinal visits of the same participant retained within one partition.

CFPs were included in FAA training when the image file could be successfully loaded, and valid chronological-age and eye-specific AMD severity annotations were available. No missing age or severity labels were imputed. The analysis used CFPs that had been prepared as $224 \times 224$ RGB images before model loading. During training, images were randomly rotated by up to $10^\circ$ and horizontally flipped with a probability of 0.5. Images were then converted to tensors and normalized using the ImageNet setting. Validation and inference images underwent the same tensor conversion and normalization without random augmentation. No additional automated image-quality score or post hoc image-quality threshold was applied beyond image availability, successful loading, and availability of the required annotations.

In the AREDS data used in our study, this information was represented using a 12-step AMD severity scale: levels 1--9 represented no/early/intermediate AMD stage, whereas levels 10--12 represented late-AMD categories (central geographic atrophy, neovascular AMD, or both). Because our prediction target was progression to any late AMD and did not aim to differentiate late-AMD subtypes, we collapsed the original levels 10--12 into a single late-AMD category before constructing the ranking loss. Specifically, for an original 12-step severity label \(s_{i}^{raw}\) we defined the ranking label \(s_{i}\) as:
\begin{equation}
    s_{i} = \min(s_{i}^{raw},10).
\end{equation}

Here, the ranking label used an effective 10-level ordinal label, comprising the original non-late AMD levels 1--9 and a unified late-AMD level 10.

The four-category grouping including no AMD (score 1), early AMD (scores 2--5), intermediate AMD (scores 6--9), and late AMD (scores 10--12) was used only for descriptive and stage-stratified analyses (\zcref{tab:cohort-characteristics, fig:faa relationship,fig:km strratified}). It was \textbf{not} used for the ranking loss or the primary Cox models. Finally, GEOAWI, RPEDWI, INCPWI, and DRARWI represent individual retinal phenotypes recorded on native 0--7 scales in AREDS. These native scales were used in the primary AREDS internal-test analyses, whereas variable-specific harmonized coding was used for the AREDS2 external-validation analyses, as described in \zcref{ssec:harmonization}. These phenotypes were used only as Cox-model covariates and were not used to define the FAA ranking labels or the four AMD stages.

For each image, the FAA age gap \(g_{i}\) was defined as:
\begin{equation}
g_{i} = {\hat{a}}_{i} - a_{i},
\end{equation}
where \({\hat{a}}_{i}\) is FAA and \(a_{i}\)is chronological age. We developed the FAA estimation model to improve prediction robustness compared to traditional convolutional neural network (CNN) methods. Unlike standard CNN architectures, the VMamba backbone is uniquely suited to effectively capture both local and long-range spatial information in retinal fundus images. By capturing long-range spatial dependencies through a state-space architecture, our VMamba-based network adaptively focuses on subtle but clinically meaningful features, such as minor microvascular alterations and small-scale drusen formations, which might be overlooked by conventional CNN approaches. This facilitates comprehensive feature extraction, enabling more accurate estimation of FAA across various stages of AMD.

Previously, regression-based age prediction models were typically trained using the mean absolute error (MAE) loss, a widely used metric in age estimation frameworks:
\begin{equation}
\mathcal{L}_{MAE} = \frac{1}{|B|}\sum_{i \in B}^{}{|{\hat{a}}_{i} - a_{i}|},
\end{equation}
where \(B\) denotes the set of images in the minibatch. However, we observed that models trained exclusively on normal (non-AMD) images significantly underestimated FAA when applied to images affected by AMD. Therefore, we only use this loss on images with no-AMD to avoid underestimating on AMD images:

\begin{equation}
\mathcal{L}_{MAE}^{N} = \frac{1}{|N|}\sum_{i \in N}^{}{|{\hat{a}}_{i} - a_{i}|},
\end{equation}
where \(N\) denotes the set of no-AMD images in the minibatch \(B\).

Besides, we introduced a ranking-based loss function to explicitly guide the model in capturing age increments associated with increasing AMD severity. Specifically, for any two images \((i,j)\) within a training batch, with corresponding chronological ages \((a_{i},a_{j})\), predicted AMD age \(({\hat{a}}_{i},{\hat{a}}_{j})\), and ranking label \((s_{i},s_{j})\), we apply a pairwise ranking constraint to encourage the model to capture progressive AMD-related aging patterns across disease stages. Specifically, for any two samples \(i\) and \(j\), if \(i\) has more severe AMD than \(j\ \) (\(s_{i} > s_{j}\)), we require \(g_{i} > g_{j}\), so that the relative ordering of FAA reflects disease severity. Inspired by the formulation of the Cox proportional hazards loss (CoxPH), we defined our ranking loss as:
\begin{equation}
L_{rank}\left(\hat{a}, a, s\right) = - \frac{1}{|B|}\sum_{i = 1}^{B}\left(g_{i} - \log\sum_{s_{i} > s_{j}}^{j}{\exp g_{j}} \right).
\end{equation}

This ranking loss explicitly incorporates AMD severity scores, allowing the model to generalize the full spectrum of AMD presentations and significantly improve the predictive utility of estimated FAA in AMD risk assessment. The final objective was:
\begin{equation}
\mathcal{L}_{total} = \mathcal{L}_{MAE}^{N} + \lambda L_{rank}.
\end{equation}

where \(\lambda\) controls the contribution of the ranking loss relative to the MAE loss and was set to 2 in all experiments.

The FAA model used \emph{VMamba-small-s2l15} as the image backbone. The network was initialized from the ImageNet pretrained checkpoint. The original classification head was replaced by a linear layer producing one continuous FAA estimate. All backbone parameters and the newly initialized output head were fine-tuned end-to-end. Training used the Adam optimizer with a learning rate of $1 \times 10^{-4}$, a batch size of 96, and 100 epochs. Random seeds for Python, NumPy, and PyTorch were set to 42. Training was implemented in PyTorch 2.6.0 and performed on an NVIDIA A100 GPU with 80 GB of memory.

All available longitudinal images from the AREDS development participants were used as separate image-level observations during FAA training. For downstream Cox-model development and evaluation, only baseline qualifying-visit CFPs were used as model inputs.

After training, FAA for each retained baseline image was calculated by averaging the predictions from five trained FAA models. This ensemble procedure was used for both the AREDS development eyes included in Cox-model fitting and the eyes in the held-out evaluation cohorts. Consequently, the FAA values used for Cox-model development were ensemble predictions, while all values from participants in the AREDS internal-test and AREDS2 cohorts were unseen by every FAA model included in the ensemble during training. Records missing FAA, chronological age, survival time, event status, smoking status, or a phenotype required for the corresponding downstream Cox model were excluded using complete-case analysis, and no imputation was performed.

\subsection{Statistical analysis}

The performance of the survival models was evaluated using the concordance index (C-index), time-dependent area under the receiver-operating-characteristic curve (AUC), and Brier score. The C-index quantified overall discriminative ability, with higher values indicating better concordance between predicted risk and observed progression times. Time-dependent AUCs were evaluated annually from years 2--7 in AREDS and from years 2--5 in AREDS2. Because follow-up duration differed between the two cohorts, the Brier score was evaluated at 5 years in AREDS and at 2 years in AREDS2. Lower Brier scores indicated lower prediction error.

Time-dependent AUCs and Brier scores were estimated using inverse-probability-of-censoring weighting, as implemented by the \verb|cumulative_dynamic_auc| and \verb|brier_score| functions in scikit-survival. For held-out evaluation in AREDS and AREDS2, the censoring distribution was estimated from the AREDS development cohort using the Kaplan--Meier estimator. IPCW was used specifically for the time-dependent AUC and Brier score and relied on the independent-censoring assumption.

For performance comparisons in \zcref{fig:prognostic performance,fig:prognostic performance human,fig:against benchmarks,fig:ablation}, 95\% confidence intervals and two-sided paired P values were estimated using participant-level bootstrap resampling with 5,000 iterations. In each iteration, participants were sampled with replacement, and all eligible eyes from each sampled participant were retained together. The same participant-level bootstrap sample was used for the models being compared in each iteration. Because the C-index, time-dependent AUC, and Brier score measure related aspects of predictive performance within the same analytical samples, no formal multiplicity correction was applied. Significance markers are defined as follows: * $p < 0.05$, ** $p < 0.01$, *** $p < 0.001$.

Kaplan--Meier curves were constructed at the eye level to visualize relative risk stratification in the held-out AREDS internal-test cohort. All stratification thresholds were determined exclusively from the AREDS development cohort before evaluation in the internal-test cohort. Three prespecified thresholding schemes were used for different analytical purposes. First, for the overall risk-stratification analyses in \zcref{fig:km}, the cutoff was defined as the median predicted Cox risk among all eligible eyes in the AREDS development cohort. Separate development-derived medians were calculated for the chronological-age-based and FAA-based Cox models because their predicted-risk scales differed.

Second, for the subgroup analyses in \zcref{fig:km others}, the fitted Cox model was not re-estimated within each subgroup. Instead, after applying the same prespecified subgroup definition to the development and internal-test cohorts, the cutoff was defined as the median predicted risk among development eyes satisfying that subgroup criterion. The resulting subgroup-specific development cutoff was then applied unchanged to internal-test eyes satisfying the same criterion. The evaluated subgroups included eyes from participants aged $\ge 70$ years, eyes with an FAA age gap of $\ge 10$ years, and eyes from participants with a smoking history.

Third, for the stage-stratified FAA analyses in \zcref{fig:km strratified}, a separate cutoff was defined as the median FAA among development eyes within each baseline AMD stage. The corresponding stage-specific FAA cutoff was then applied unchanged to internal-test eyes within the same stage.

These median splits were used to visualize relative risk separation within the prespecified population or stratum. They were not optimized clinical decision thresholds and should not be interpreted as universally transportable or directly actionable cutoffs. Group differences in the Kaplan--Meier analyses were assessed using two-sided log-rank tests. Hazard ratios and 95\% confidence intervals for the high-versus-low comparisons, where reported, were estimated separately using Cox proportional-hazards models. Accordingly, the displayed log-rank P values were obtained independently of the Cox-model hazard ratios and confidence intervals. Eye-level sample sizes and late-AMD progression event counts for the overall, subgroup-specific, and stage-stratified Kaplan--Meier analyses are reported in \zcref{tab:supp-risk-events}.

Statistical analyses were conducted using Python 3.10, Lifelines 0.30.0, scikit-survival 0.23.1, and scikit-learn 1.5.2.

\paragraph{Ethics approval.}

This study's data analysis was reviewed and approved by the Institutional Review Board of Weill Cornell Medicine (IRB Approval No. 21-08023836).

\paragraph{Data availability.}

The individual-level clinical, phenotypic, and color fundus photograph data analyzed in this study were obtained from the controlled-access National Institutes of Health resources for the Age-Related Eye Disease Study (AREDS) and Age-Related Eye Disease Study 2 (AREDS2). The AREDS data set (NCT00000145) generated during and/or analyzed during the current study is available in the dbGAP repository (\url{https://www.ncbi.nlm.nih.gov/projects/gap/cgi-bin/study.cgi?study_id=phs000001.v3.p1}). The AREDS2 data set generated during and/or analyzed during the current study has been deposited in dbGAP, accession number phs002015.v1.p1. However, due to an ongoing IRB review of the AREDS2 data set, the data will not be available in dbGAP for $\sim 3$ months post-publication. While the AREDS2 data set is being reviewed by the IRB, it is available from the corresponding author upon reasonable request.

\paragraph{Code availability.}\label{code-availability}

Source code for FAA model training and inference, Cox-model development and evaluation, statistical analyses, and figure generation is publicly available through the FundusAge GitHub repository at \url{https://github.com/bionlplab/FundusAge}. The analyses were implemented using Python 3.10, PyTorch 2.6.0, Lifelines 0.30.0, scikit-survival 0.23.1, and scikit-learn 1.5.2. The repository does not contain individual-level AREDS or AREDS2 clinical data or color fundus photographs. Researchers seeking to reproduce the analyses must obtain independent authorized access to the source datasets through dbGaP and configure the corresponding local data paths as described in the repository documentation.

\paragraph{Acknowledgments.}\label{acknowledgements}

This work was supported by the National Eye Institute grant R21EY035296 and R01EY037701, the NIH Intramural Research Program at the National Library of Medicine, and the Intramural Research Program at the National Eye Institute. The funder had no role in the design and conduct of the study; collection, management, analysis, and interpretation of the data; preparation, review, or approval of the manuscript; and decision to submit the manuscript for publication.

\paragraph{Author Contributions.}\label{author-contributions}

Study concepts/study design, H.D., Y.P., Z.L.; clinical expertise and data interpretation, E.C.; supervision, Y.P., Z.L.; manuscript drafting, H.D.; manuscript revision for important intellectual content, H.D., B.H., P.W., M.L., E.C., Z.L., Y.P.; approval of final version of the submitted manuscript, all authors; agrees to ensure any questions related to the work are appropriately resolved, all authors; literature research, H.D.; experimental studies, H.D., B.H., P.W.; data interpretation and statistical analysis, H.D., B.H., Y.P.; and manuscript editing, all authors.

\paragraph{Competing Interests.}\label{competing-interests}

The authors declare no competing financial or non-financial interests.

\setlength{\bibsep}{3pt plus 0.3ex}
\bibliographystyle{unsrtnat}
\bibliography{ref}

\newpage
\appendix

\setcounter{table}{0}
\renewcommand\tablename{eTable}

\setcounter{figure}{0}
\renewcommand\figurename{eFigure}

\zcsetup{
    countertype={
        table=etable,
        figure=efigure,
        section=esection,
        subsection=esection
    }
}

\part*{Supplementary materials}
\label{sec:appendix}

\section{Fold-specific and ensemble downstream prognostic performance of the FAA-based Cox models}

To assess the consistency of downstream prognostic performance across the five FAA models, we evaluated Cox models using FAA estimates generated by each fold-specific model and by the five-model ensemble. \textbf{\zcref{tab:supp-performance}} reports the C-index, 5-year time-dependent AUC, and 5-year Brier score under the reading-center-defined and specialist-defined phenotype settings. These metrics characterize downstream prognostic performance rather than direct FAA age-estimation accuracy.

\begin{table}[H]
\centering
\caption{Fold-specific and ensemble downstream prognostic performance of the FAA-based Cox models under the reading-center-defined and specialist-defined phenotype settings.}
\label{tab:supp-performance}
\small
\begin{tabular}{lcccccc}
\toprule
& \multicolumn{3}{l}{Reading-center-defined setting}
& \multicolumn{3}{l}{Specialist-defined setting} \\
\cmidrule(lr){2-4}
\cmidrule(lr){5-7}
Fold & C-index & 5y-AUC & 5y-Brier & C-index & 5y-AUC & 5y-Brier \\
\midrule
1   & 0.878 & 0.878 & 0.048 & 0.842 & 0.850 & 0.051 \\
2   & 0.869 & 0.876 & 0.048 & 0.842 & 0.852 & 0.051 \\
3   & 0.883 & 0.887 & 0.047 & 0.849 & 0.860 & 0.050 \\
4   & 0.886 & 0.882 & 0.047 & 0.851 & 0.858 & 0.051 \\
5   & 0.882 & 0.888 & 0.047 & 0.847 & 0.859 & 0.049 \\
\midrule
Ensemble & 0.888 & 0.888 & 0.046 & 0.857 & 0.861 & 0.049 \\
\bottomrule
\end{tabular}
\end{table}

\newpage

\section{Definition of AMD Phenotype Scores}
\label{ssec:amd definition}

In our analysis, AMD phenotypes were graded according to standardized scoring scales with defined ranges. Each phenotype's severity score was assigned based on specific clinical criteria derived from retinal fundus images. The definitions of scoring values for each phenotype are provided as follows:

\subsection{AMD Phenotypes in AREDS}

\begin{enumerate}[nosep]
\item \textbf{GEOAWI}: Geographic atrophy within grid

\begin{enumerate}[nosep, label=\arabic*:, start=0]
\item None
\item Questionable
\item Present, $<$ Circle I-2 (diameter, 0.058 disc area)
\item Present, $\ge$ Circle I-2, $<$ Circle O-2 (diameter, 0.19 disc area)
\item Present, $\ge$ Circle O-2, $<$ 1/2 disc area
\item Present, $\ge$ 1/2 disc area, $<$ 1 disc area
\item Present, $\ge$ 1 disc area, $<$ 2 disc areas
\item Present, $\ge$ 2 disc areas
\end{enumerate}

\item \textbf{RPEDWI}: Rpe depigmentation area w/i grid

\begin{enumerate}[nosep, label=\arabic*:, start=0]
\item None
\item Questionable
\item Present, $<$ Circle I-2 (diameter, 0.058 disc area)
\item Present, $\ge$ Circle I-2, $<$ Circle O-2 (diameter, 0.19 disc area)
\item Present, $\ge$ Circle O-2, $<$ 1/2 disc area
\item Present, $\ge$ 1/2 disc area, $<$ 1 disc area
\item Present, $\ge$ 1 disc area, $<$ 2 disc areas
\item Present, $\ge$ 2 disc areas
\end{enumerate}

\item \textbf{INCPWI}: Increased pigment area w/i grid

\begin{enumerate}[nosep, label=\arabic*:, start=0]
\item None
\item Questionable
\item Present, $<$ Circle C0 (diameter, 0.0017 disc area)
\item Present, $\ge$ Circle C0, $<$ Circle C1 (diameter, 0.0069 disc area)
\item Present, $\ge$ Circle C0, $<$ Circle C2 (diameter, 0.028 disc area)
\item Present, $\ge$ Circle C2, $<$ Circle O2 (diameter, 0.19 disc area)
\item Present, $\ge$ Circle O2
\item Other
\end{enumerate}

\item \textbf{DRARWI}: Drusen area w/i grid

\begin{enumerate}[nosep, label=\arabic*:, start=0]
\item None/Questionable/less than Circle C0 (diameter, 0.0017 disc area)
\item Present, $\ge$ Circle C0, $<$ Circle C1 (diameter, 0.0069 disc area)
\item Present, $\ge$ Circle C0, $<$ Circle C2 (diameter, 0.028 disc area)
\item Present, $\ge$ Circle C2, $<$ Circle I2 (diameter, 0.058 disc area)
\item Present, $\ge$ Circle I2, $<$ Circle O2 (diameter, 0.058 disc area)
\item Present, $\ge$ Circle O-2, $<$ 1/2 disc area
\item Present, $\ge$ 1/2 disc area, $<$ 1 disc area
\item Present, $\ge$ 1 disc areas
\end{enumerate}

\item \textbf{DRUSENQ}: Clinical drusen size assessment

\begin{enumerate}[nosep, label=\arabic*:, start=0]
\item None or small drusen
\item Intermediate drusen
\item Large drusen (125 U or greater)
\item Advanced AMD
\end{enumerate}

\item \textbf{PIGDRUQ}: Pigmentary abnormalities

\begin{enumerate}[nosep, label=\arabic*:, start=0]
\item No
\item Yes
\end{enumerate}
\end{enumerate}

\subsection{AMD Phenotypes in AREDS2}

\begin{enumerate}[nosep]
\item \textbf{DPINGRID}: Depigmentation within center and inner subfields

\begin{enumerate}[nosep, label=\arabic*:, start=0]
\item Absent
\item Questionable
\item Present, $<$ Circle I2 (diameter, 0.058 disc area)
\item Present, $\ge$ Circle I2, $<$ 1/2 disc area
\item Present, $\ge$ 1/2 disc area
\end{enumerate}

\item \textbf{IPINGRID}: Increased pigment within center and inner subfields

\begin{enumerate}[nosep, label=\arabic*:, start=0]
\item Absent
\item Questionable
\item Present, $<$ Circle C2 (diameter, 0.028 disc area)
\item Present, $\ge$ Circle C2
\end{enumerate}

\item \textbf{DARINGRD}: Drusen area within grid (same as \textbf{DRARWI} in Areds)

\begin{enumerate}[nosep, label=\arabic*:, start=0]
\item None/Questionable/less than Circle C0 (diameter, 0.0017 disc area)
\item Present, $\ge$ Circle C0, $<$ Circle C1 (diameter, 0.0069 disc area)
\item Present, $\ge$ Circle C0, $<$ Circle C2 (diameter, 0.028 disc area)
\item Present, $\ge$ Circle C2, $<$ Circle I2 (diameter, 0.058 disc area)
\item Present, $\ge$ Circle I2, $<$ Circle O2 (diameter, 0.058 disc area)
\item Present, $\ge$ Circle O-2, $<$ 1/2 disc area
\item Present, $\ge$ 1/2 disc area, $<$ 1 disc area
\item Present, $\ge$ 1 disc areas
\end{enumerate}

\item \textbf{GADAGRID}: Geographic atrophy, area within grid in disc area(s) (DA): 0.00-16.00 DA

This variable was recoded to align with the \textbf{GEOAWI} scale used in AREDS.

\item \textbf{EXAM}: Drusen status (same as \textbf{DRUSENQ} in AREDS)

\begin{enumerate}[nosep, label=\arabic*:, start=0]
\item None or small drusen
\item Intermediate drusen
\item Large drusen (125 U or greater)
\item Advanced AMD
\end{enumerate}

\item \textbf{PIGAB}: Pigmentary abnormalities (same as \textbf{PIGDRUQ} in AREDS)

\begin{enumerate}[nosep, label=\arabic*:, start=0]
\item No
\item Yes
\end{enumerate}
\end{enumerate}

\subsection{Harmonization of phenotype variables between AREDS and AREDS2}
\label{ssec:harmonization}

Phenotype harmonization was used specifically for the AREDS2 external-validation analysis. The primary AREDS internal-test Cox models were fitted and evaluated using the native AREDS phenotype coding. For external validation, corresponding variables were mapped to common, variable-specific numerical scales, and the harmonized Cox models were fitted using the recoded AREDS development data before application to AREDS2. No AREDS2 participant-level outcomes were used for model fitting, recalibration, model selection, or threshold determination.

For the RPEDWI/DPINGRID variable pair, AREDS RPEDWI was coarsened from its native 0--7 scale to the DPINGRID-compatible 0--4 scale using the following mapping: 0$\rightarrow$0, 1$\rightarrow$1, 2$\rightarrow$2, 3--4$\rightarrow$3, and 5--7$\rightarrow$4. AREDS2 DPINGRID was already recorded on the corresponding 0--4 scale and was used without further transformation.

For the INCPWI/IPINGRID variable pair, AREDS INCPWI was coarsened from its native 0--7 scale to the IPINGRID-compatible 0--3 scale using the following mapping: 0$\rightarrow$0, 1$\rightarrow$1, 2--4$\rightarrow$2, and 5--7$\rightarrow$3. AREDS2 IPINGRID was already recorded on the corresponding 0--3 scale and was used without further transformation.

For geographic atrophy, the continuous AREDS2 GADAGRID variable, which records geographic-atrophy area in disc areas, was discretized according to the native AREDS GEOAWI thresholds. Values were coded as 0 for absence, 1 when the source annotation explicitly indicated questionable geographic atrophy, 2 for an area greater than 0 and less than 0.058 disc areas, 3 for 0.058 to less than 0.19 disc areas, 4 for 0.19 to less than 0.5 disc areas, 5 for 0.5 to less than 1 disc area, 6 for 1 to less than 2 disc areas, and 7 for 2 or more disc areas. AREDS GEOAWI was retained on its native 0--7 scale.

AREDS DRARWI and its AREDS2 counterpart DARINGRD already shared the same 0--7 coding and were used without transformation. Similarly, DRUSENQ/EXAM and PIGDRUQ/PIGAB shared 0--3 and binary coding, respectively.

Accordingly, the primary AREDS internal-test results were obtained using native AREDS phenotype coding, whereas the AREDS2 external-validation results were obtained using harmonized Cox models fitted solely in the AREDS development cohort and subsequently applied to AREDS2.

\newpage

\section{Statistical Analysis}

\subsection{Cox Proportional Hazards Models}

We constructed Cox proportional hazards models to evaluate the prognostic value of the Fundus-based AMD Age (FAA) and comparison baselines. All models were built using the \textbf{CoxPH implementation in the Lifelines Python package}.

For each model, the following covariates were included depending on the phenotype setting:

\begin{itemize}[nosep]
\item
  \textbf{Reading-center--defined models in AREDS dataset:}

  \begin{itemize}[nosep]
  \item
    FAA (or chronological age, depending on the comparison)
  \item
    Geographic atrophy within grid (GEOAWI)
  \item
    RPE depigmentation within grid (RPEDWI)
  \item
    Increased pigment within grid (INCPWI)
  \item
    Drusen area within grid (DRARWI)
  \item
    Smoking status (SMK)
  \end{itemize}
\item
  \textbf{Reading-center--defined models in AREDS2 dataset:}

  \begin{itemize}[nosep]
  \item
    FAA (or chronological age, depending on the comparison)
  \item
    Geographic atrophy within grid (GEOAWI)
  \item
    RPE depigmentation within grid (RPEDWI), recode to 0-4 to fit with definition in AREDS2
  \item
    Increased pigment within grid (INCPWI), recode to 0-3 to fit with definition in AREDS2
  \item
    Drusen area within grid (DRARWI)
  \item
    Smoking status (SMK)
  \end{itemize}
\item
  \textbf{Specialist-defined models in both AREDS and AREDS2:}

  \begin{itemize}[nosep]
  \item
    FAA (or chronological age)
  \item
    Clinical drusen size assessment (DRUSENQ)
  \item
    Pigment abnormalities (PIGDRUQ)
  \item
    Smoking status (SMK)
  \end{itemize}
\end{itemize}

All phenotype variables were used in their \textbf{ordinal or binary encoded form}, as described in \zcref{tab:supp-cox-variables}, which provides full details of variable definitions, coding schemes, and units. No additional composite scoring or binary recoding was performed beyond the prespecified variable harmonization described in \zcref{ssec:harmonization}.

\begin{table}[H]
\centering
\caption{Specification of variables included in the Cox proportional hazards models.}
\label{tab:supp-cox-variables}
\small
\begin{tabularx}{\textwidth}{llLL}
\toprule
Variable & Type & Unit / Coding & Description \\
\midrule
FAA & Continuous & Per 1-year increase &
Fundus-based AMD age predicted from CFP \\\grayline
Chronological age & Continuous & Per 1-year increase &
Age at baseline visit \\\grayline
AMD severity score & Ordinal & Scale 1--12 &
Baseline AMD severity level \\\grayline
Smoking status (SMK) & Categorical &
Never (reference), former, current &
Self-reported smoking history \\\grayline
GEOAWI & Ordinal & 0--7 &
Geographic atrophy within grid \\\grayline
RPEDWI & Ordinal & 0--7 &
RPE depigmentation area \\\grayline
INCPWI & Ordinal & 0--7 &
Increased pigment area \\\grayline
DRARWI & Ordinal & 0--7 &
Drusen area \\
\bottomrule
\end{tabularx}
\end{table}

\subsection{Unit of Analysis and Single-Eye Sensitivity Analyses}

All survival outcomes were defined at the eye level. Each eligible eye contributed a separate analytical record containing an eye-specific baseline color fundus photograph, FAA estimate, AMD phenotype measurements, progression indicator, and follow-up time. A participant could contribute one or two eligible eyes to the primary analyses.

To directly evaluate whether the primary findings were driven by the inclusion of correlated bilateral observations, we performed strict single-eye sensitivity analyses. Within each phenotype-definition setting, the Cox models were evaluated using only left or only right eyes from the held-out internal-test cohort. Thus, each participant contributed \textbf{at most one eye}. The eligibility criteria, covariate sets, Cox-model specifications, and penalization parameters were identical to those used in the primary analyses with no side-specific model selection or threshold optimization. C-index confidence intervals were estimated using 5,000 bootstrap resamples. Because each participant contributed at most one eye to a given side-specific dataset, these sensitivity analyses did not contain correlated fellow-eye observations.

In the reading-center-defined setting (\zcref{tab:supp-single-eye}), the C-index increased from 0.863 (95\% CI, 0.807--0.911) to 0.908 (95\% CI, 0.870--0.941) in the left-eye-only analysis, and from 0.805 (95\% CI, 0.731--0.864) to 0.865 (95\% CI, 0.812--0.910) in the right-eye-only analysis when chronological age was replaced with FAA. Similar improvements were observed in the specialist-defined setting. In the left-eye-only analysis, the C-index increased from 0.824 (95\% CI, 0.773--0.873) to 0.882 (95\% CI, 0.843--0.902). In the right-eye-only analysis, the C-index increased from 0.790 (95\% CI, 0.725--0.852) to 0.834 (95\% CI, 0.779--0.886).

Across all four single-eye analyses, FAA consistently improved the C-index by 0.044 to 0.060 relative to chronological age. While the confidence intervals were wider because of the smaller sample sizes, the improvements were consistent with the primary analyses. These results indicate that the observed benefit of FAA was not driven by the inclusion of correlated bilateral eyes.

\begin{table}[H]
\centering
\caption{Single-eye sensitivity analyses of prognostic discrimination in the AREDS testing cohort. Values are C-indices with 95\% confidence intervals. $N$ denotes the number of eligible eyes in the AREDS testing cohort. In the left-eye-only and right-eye-only analyses, each participant contributed at most one eye. The both-eye results are shown for reference.}
\label{tab:supp-single-eye}
\small
\begin{tabular}{lcc}
\toprule
& Chronological age-based model & FAA-based model \\
\midrule
\rowcolor{gray!20}\textit{Reading center-based model} && \\
Both eyes ($N=769$)  & 0.836 (0.788--0.880) & 0.888 (0.853--0.919) \\
Left eye ($N=394$)   & 0.863 (0.807--0.911) & 0.908 (0.870--0.941) \\
Right eye ($N=375$)  & 0.805 (0.731--0.864) & 0.865 (0.812--0.910) \\
\midrule
\rowcolor{gray!20}\textit{Specialist-defined model}&& \\
Both eyes ($N=811$)  & 0.808 (0.761--0.851) & 0.857 (0.817--0.892) \\
Left eye ($N=408$)   & 0.824 (0.773--0.873) & 0.882 (0.843--0.902) \\
Right eye ($N=403$)  & 0.790 (0.725--0.852) & 0.834 (0.779--0.886) \\
\bottomrule
\end{tabular}
\end{table}

\subsection{Assessment of Cox Proportional-Hazards Assumptions and Covariate Functional Forms}
\label{ssec:assessment-of-cox-proportional-hazards-assumptions-and-covariate-functional-forms}

The following analyses were performed in the AREDS development cohort after the primary Cox-model specifications had been fixed. Separate diagnostic analyses were conducted for the reading-center-defined and specialist-defined models containing either chronological age or fundus-based AMD age (FAA). Within each phenotype-definition setting, the chronological-age-based and FAA-based models used the same complete-case analytical samples and the same smoking and ocular phenotype covariates; only the age representation differed.

\subsubsection{Proportional-hazards assumption}
\label{ssec:proportional-hazards-assumption}

The proportional-hazards assumption was assessed using scaled Schoenfeld residuals. Covariate-specific tests were performed for each primary Cox model. A two-sided P value below 0.05 was considered evidence against the proportional-hazards assumption. These diagnostics were used to assess the stability of covariate effects over follow-up and were not used for data-driven variable selection.

There was no clear evidence of nonproportional hazards for chronological age or FAA in either phenotype-definition setting (\zcref{tab:supp-ph-assumption}). In the reading-center-defined analyses, the covariate-specific p-values were 0.434 for chronological age and 0.050 for FAA. One isolated signal was observed for DRARWI in the chronological-age-based model ($p=0.018$); all other covariates in the reading-center FAA model had $p>0.05$. In the specialist-defined analyses, all covariate-specific tests had $p>0.05$, including chronological age ($p=0.532$) and FAA ($p=0.190$).

\begin{table}[H]
\centering
\caption{Covariate-specific scaled Schoenfeld-residual tests for the proportional-hazards assumption in the AREDS development cohort. P-values are two-sided and unadjusted. A P-value below 0.05 indicates evidence against the proportional-hazards assumption. DRARWI, drusen area within the grid; GEOAWI, geographic atrophy within the grid; INCPWI, increased pigment area within the grid; RPEDWI, retinal pigment epithelium depigmentation area within the grid; DRUSENQ, clinical drusen size assessment; PIGDRUQ, pigmentary abnormalities.}
\label{tab:supp-ph-assumption}
\small
\begin{tabular}{llrr}
\toprule
Age representation & Variable tested & $\chi^2$ & p-value \\
\midrule
\rowcolor{gray!20}\multicolumn{4}{l}{\textbf{A. Reading center}} \\
Chronological age & Chronological age & 0.613 & 0.434 \\
& DRARWI           & 5.550 & 0.018 \\
& GEOAWI           & 2.019 & 0.155 \\
& INCPWI           & 0.417 & 0.519 \\
& RPEDWI           & 3.237 & 0.072 \\
& Smoking history  & 0.025 & 0.874 \\
\midrule
FAA & FAA              & 3.831 & 0.050 \\
& DRARWI           & 1.259 & 0.262 \\
& GEOAWI           & 0.084 & 0.772 \\
& INCPWI           & $<0.001$ & 0.998 \\
& RPEDWI           & 1.364 & 0.243 \\
& Smoking history  & 0.449 & 0.503 \\
\midrule
\rowcolor{gray!20}\multicolumn{4}{l}{\textbf{B. Specialist}}\\
Chronological age & Chronological age & 0.391 & 0.532 \\
& DRUSENQ           & 2.975 & 0.085 \\
& PIGDRUQ           & 0.591 & 0.442 \\
& Smoking history   & 1.001 & 0.317 \\
\midrule
FAA & FAA              & 1.719 & 0.190 \\
& DRUSENQ          & 3.161 & 0.075 \\
& PIGDRUQ          & 0.303 & 0.582 \\
& Smoking history  & 0.315 & 0.575 \\
\bottomrule
\end{tabular}
\end{table}

\subsubsection{Functional form of FAA and chronological age}
\label{ssec:functional-form-of-faa-and-chronological-age}

Potential nonlinear associations of FAA and chronological age with the log hazard were evaluated using restricted cubic splines (\zcref{tab:supp-spline}). Four knots were placed at the 5th, 35th, 65th, and 95th percentiles of the corresponding variable in the AREDS development cohort. For each age representation, an unpenalized spline model was compared with an otherwise identical unpenalized model containing a single linear term by likelihood-ratio testing. The held-out AREDS internal-test C-index was also calculated for both specifications to evaluate whether a statistically detectable departure from linearity materially changed discrimination.

In the reading-center-defined setting, departures from strict linearity were detected for chronological age ($p=0.034$) and FAA ($p<0.001$). In the specialist-defined setting, no evidence of nonlinearity was observed for chronological age ($p=0.224$), whereas a modest nonlinear association was detected for FAA ($p=0.025$). However, the spline specifications changed held-out C-indices by no more than 0.005. The FAA-based model retained higher discrimination than the chronological-age-based model under the spline specification in both phenotype-definition settings. The prespecified linear terms were therefore retained in the primary prediction models for parsimony and direct comparability, while the corresponding per-year hazard ratios are interpreted as average log-linear associations over the observed range rather than constant effects at every predictor value.

\begin{table}[H]
\centering
\caption{Restricted-cubic-spline assessment of the functional forms of chronological age and FAA. The likelihood-ratio test compares an unpenalized restricted cubic spline with an otherwise identical unpenalized model containing a single linear term. Delta C-index is spline minus linear. Knots were placed at the 5th, 35th, 65th, and 95th percentiles. Knot values were 61, 66, 70, and 77 years for chronological age in both settings; 65.39, 70.01, 74.69, and 83.58 years for FAA in the reading-center setting; and 65.35, 69.93, 74.56, and 83.51 years for FAA in the specialist setting.}
\label{tab:supp-spline}
\small
\begin{tabular}{lrrrrr}
\toprule
Age representation
& LRT $\chi^2$ (df)
& \makecell[r]{$P$ for \\nonlinearity}
& \makecell[r]{Partial AIC, \\linear / spline}
& \makecell[r]{Test C-index, \\linear / spline}
& $\Delta$ C-index \\
\midrule
\rowcolor{gray!20}\multicolumn{6}{l}{\textbf{A. Reading center}}\\
Chronological age
& 6.764 (2)
& 0.034
& 6423.17 / 6420.40
& 0.837 / 0.842
& +0.0049 \\
FAA
& 32.618 (2)
& $<0.001$
& 6179.97 / 6151.35
& 0.888 / 0.887
& -0.0007 \\
\rowcolor{gray!20}\multicolumn{6}{l}{\textbf{B. Specialist}}\\
Chronological age
& 2.992 (2)
& 0.224
& 6239.52 / 6240.53
& 0.808 / 0.810
& +0.0015 \\
FAA
& 7.383 (2)
& 0.025
& 6008.44 / 6005.06
& 0.856 / 0.862
& +0.0051 \\
\bottomrule
\end{tabular}
\end{table}

\begin{figure}[H]
\centering
\includegraphics[width=\textwidth]{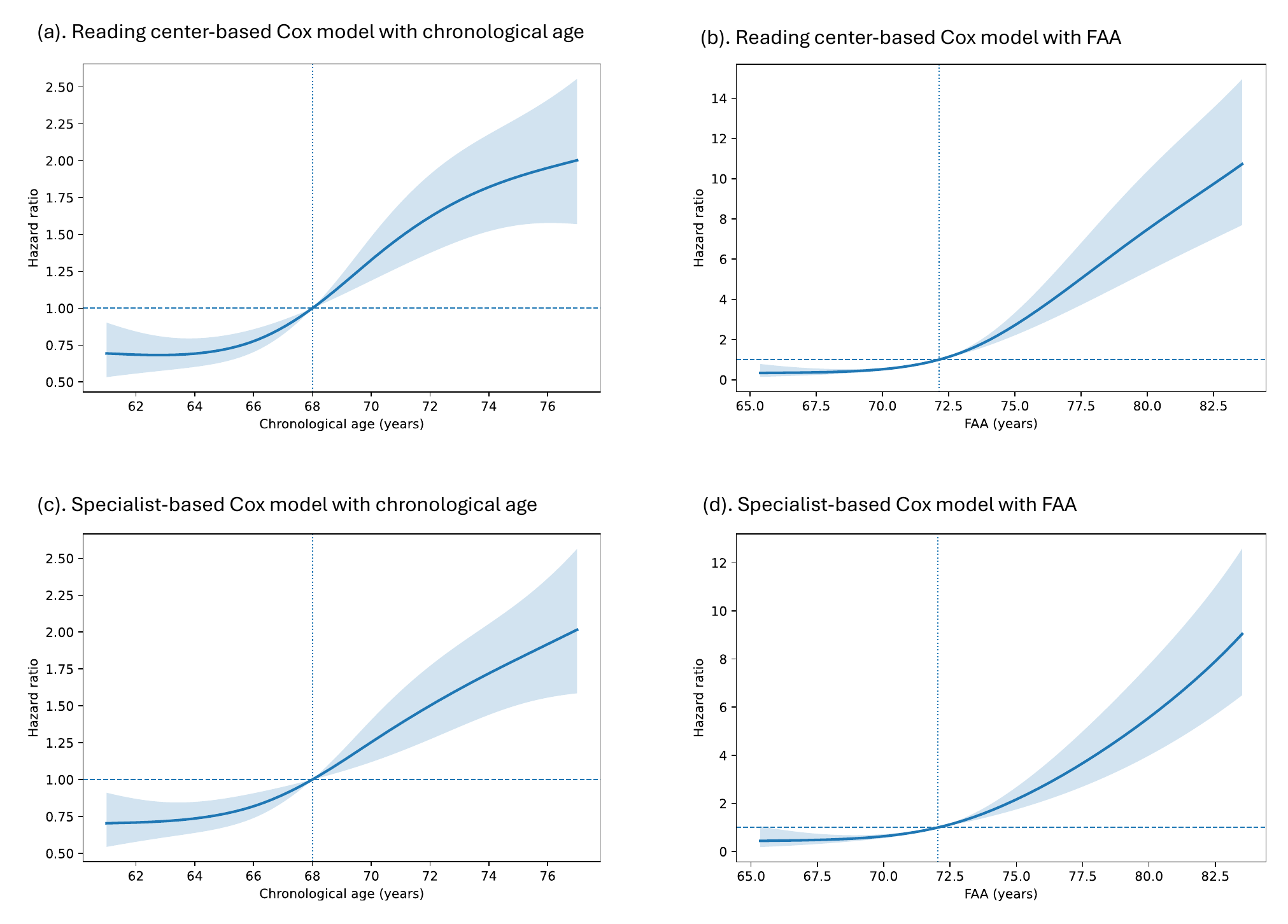}

\caption{\textbf{Restricted-cubic-spline associations of chronological age and FAA with late-AMD progression.} \textbf{a-b}, Reading-center-based models containing (a) chronological age or (b) FAA. \textbf{c-d}, Specialist-based models containing (c) chronological age or (d) FAA. Solid lines show the estimated hazard ratios and shaded regions show 95\% confidence intervals, relative to the development-cohort median of the corresponding age representation. Curves are displayed from the 5th to the 95th percentile; the horizontal dashed line indicates a hazard ratio of 1 and the vertical dotted line indicates the reference median. Knots were placed at the 5th, 35th, 65th, and 95th percentiles.}
\end{figure}

\subsubsection{Sensitivity to categorical coding of ordinal phenotype grades}
\label{ssec:sensitivity-to-categorical-coding-of-ordinal-phenotype-grades}

In the primary models, ordered phenotype grades were entered as linear score terms, such that each hazard ratio represented an average change in log hazard per one-grade increase. To evaluate whether the FAA-versus-chronological-age comparison depended on the assumption of equal risk increments between adjacent grades, the nonbinary phenotype variables were reparametrized as categorical indicator variables. The chronological-age-based and FAA-based models were then refitted using the same complete-case analytical samples, covariates, and penalization parameters. C-index confidence intervals and paired differences were estimated using 5,000 participant-level bootstrap resamples, retaining all eligible eyes from a sampled participant together.

The principal comparison was preserved under categorical phenotype coding (\zcref{tab:supp-phenotype-coding}). In the reading-center-defined setting, the FAA-based model achieved a C-index of 0.871 compared with 0.813 for the chronological-age-based model, yielding a difference of 0.058 (95\% CI, 0.026-0.093; $p<0.01$). In the specialist-defined setting, the corresponding C-indices were 0.850 and 0.797, with a difference of 0.052 (95\% CI, 0.031-0.075; $p<0.01$). The direction and magnitude of the FAA-associated improvement were similar to those obtained with ordinal-linear coding, indicating that the primary conclusion was not dependent on treating adjacent phenotype grades as equally spaced risk increments.

\begin{table*}[H]
\centering
\caption{Sensitivity of prognostic discrimination to categorical rather than ordinal-linear coding of phenotype grades. Confidence intervals and P values were obtained using 5,000 participant-level paired bootstrap resamples.}
\label{tab:supp-phenotype-coding}
\small
\begin{tabular}{lrr}
\toprule
Phenotype coding
& \makecell[r]{Chronological-age model\\C-index (95\% CI)}
& \makecell[r]{FAA model C-index\\(95\% CI)} \\
\midrule
\rowcolor{gray!20}\multicolumn{3}{l}{\textbf{A. Reading center}}\\
Ordinal-linear
& 0.836 (0.788--0.880)
& 0.888 (0.853--0.919) \\
Categorical
& 0.813 (0.763--0.862)
& 0.871 (0.832--0.908) \\
\midrule
\rowcolor{gray!20}\multicolumn{3}{l}{\textbf{B. Specialist}}\\
Ordinal-linear
& 0.808 (0.761--0.851)
& 0.857 (0.817--0.892) \\
Categorical
& 0.797 (0.750--0.838)
& 0.850 (0.810--0.888) \\
\bottomrule
\end{tabular}
\end{table*}

\subsubsection{Summary of assumption and sensitivity analyses}
\label{ssec:summary-of-assumption-and-sensitivity-analyses}

Overall, the diagnostics showed no clear evidence that the proportional-hazards assumption was violated for chronological age or FAA, although an isolated covariate-specific signal was observed for DRARWI in the reading-center chronological-age model. Restricted cubic splines detected statistically significant nonlinearity in some model settings, but allowing nonlinear age functions changed held-out C-indices by no more than 0.005 and did not alter the higher discrimination of FAA-based models. Likewise, the FAA-associated improvement was retained when ordinal phenotype scores were represented categorically. These analyses support the robustness of the primary FAA-versus-chronological-age comparison while motivating cautious interpretation of single per-year and per-grade hazard ratios as average associations across the observed ranges.

\subsection{Decomposition of the Prognostic Contributions of FAA, Chronological Age, and Baseline Phenotypes}

To distinguish the prognostic information contributed by chronological age, fundus-based AMD age (FAA), smoking status, and baseline AMD phenotypes, we performed a prespecified series of Cox-model comparisons in the reading-center-defined AREDS analysis (\zcref{tab:model-comparison}). Six models were evaluated: (1) smoking status and phenotype covariates; (2) chronological age and smoking status; (3) FAA and smoking status; (4) chronological age, smoking status, and phenotype covariates; (5) FAA, smoking status, and phenotype covariates; and (6) chronological age, FAA, smoking status, and phenotype covariates. No model-specific variable screening or hyperparameter selection was performed. Discrimination was assessed using the C-index in the held-out AREDS internal-test cohort, and 95\% confidence intervals were estimated using participant-level bootstrap resampling.

Overall, the decomposition analysis showed that both baseline AMD phenotypes and FAA contributed prognostic information. FAA provided substantially higher discrimination than chronological age when each was combined with smoking status, and the combination of FAA, smoking status, and baseline phenotypes produced the highest C-index among the evaluated model specifications.

\begin{table}[H]
\centering
\caption{Decomposition of the prognostic contributions of chronological age, FAA, smoking status, and reading-center-defined AMD phenotypes in the AREDS internal-test cohort.}
\label{tab:model-comparison}
\small
\begin{tabular}{clr}
\toprule
Model & Covariates & C-index (95\% CI) \\
\midrule
1 & Smoking status + phenotypes
  & 0.811 (0.765--0.857) \\
2 & Chronological age + smoking status
  & 0.694 (0.617--0.765) \\
3 & FAA + smoking status
  & 0.869 (0.830--0.905) \\
4 & Chronological age + smoking status + phenotypes
  & 0.836 (0.788--0.880) \\
5 & FAA + smoking status + phenotypes
  & 0.888 (0.853--0.919) \\
6 & Chronological age + FAA + smoking status + phenotypes
  & 0.856 (0.810--0.896) \\
\bottomrule
\end{tabular}
\end{table}

\subsection{Adapted image-only benchmark based on Yan et al.}
\label{ssec:benchmark}

To provide a matched image-based benchmark, we adapted the publicly available implementation of Yan et al.\(^{24}\) The genetic-input branch was removed, and the adapted model used only baseline color fundus photographs as input. The benchmark was trained using the same participant-level AREDS development partition and evaluated using the same eligible AREDS internal-test eyes as the FAA-based analysis. Images from both eyes and all longitudinal visits of each participant were retained within the same data partition. The benchmark predicted progression to late AMD within 5 years from the baseline image. Late AMD was defined as the first AREDS severity score of 10--12. Eyes that progressed to late AMD within 5 years were labeled as events, whereas eyes remaining free of late AMD through 5 years were labeled as non-events. The continuous predicted probability was used as the benchmark risk score for C-index evaluation. No AREDS internal-test participant was used for model training, model selection, or hyperparameter optimization.

\newpage

\section{AMD Severity Representations Used in the Study}
\label{ssec:amd-severity-representations-used-in-the-study}

Several related but distinct representations of AMD severity were used for different analytical purposes in this study (\zcref{tab:supp-amd-severity}). The source AREDS annotations used a 12-level coding, with scores 1--9 representing non-late AMD severity and scores 10--12 representing late-AMD categories. For FAA training, scores 10--12 were combined into a single late-AMD rank of 10 because the model did not distinguish among late-AMD subtypes. For the severity-score Cox comparator, only scores 1--9 were included because eyes with late AMD at baseline had already reached the study endpoint and were excluded from survival modeling. The four-stage grouping was used only for descriptive and stage-stratified analyses.

\begin{table}[H]
\centering
\caption{Definitions and analytical roles of AMD severity representations.}
\label{tab:supp-amd-severity}
\small
\begin{tabularx}{\linewidth}{>{\hsize=0.5\hsize}Lll>{\hsize=1.5\hsize}L}
\toprule
Variable & Type & Unit/coding & Description \\
\midrule
AMD severity score
& Ordinal
& 1--12
& Eye- and visit-specific source coding. Levels 1--9 represent ordered non-late AMD severity; levels 10--12 denote late-AMD categories. Used to define baseline AMD status and the late-AMD progression endpoint. \\\grayline

FAA ranking severity label
& Ordinal
& 1--10
& Derived from original scores 1--12 by mapping scores 10--12 to a unified late-AMD rank of 10; used only in FAA training. \\\grayline

Four-stage AMD grouping
& Categorical
& 1 / 2--5 / 6--9 / 10--12
& Used only for descriptive summaries and stage-stratified analyses. \\
\bottomrule
\end{tabularx}
\end{table}

\newpage

\section{Year-specific Time-dependent AUC Analyses}
\label{ssec:year-specific}

To provide a transparent assessment of discrimination across follow-up and avoid ambiguity associated with reporting only AUC ranges, we report the complete year-specific time-dependent AUC values for both the reading-center-defined and specialist-defined analyses. Time-dependent AUCs were evaluated from years 2--7 in AREDS and from years 2--5 in AREDS2, reflecting the different follow-up durations of the two cohorts.

Across both phenotype-definition settings, the FAA-based model achieved higher time-dependent AUC point estimates than the chronological-age-based model at every evaluated horizon in both AREDS and AREDS2. Detailed results are provided in \zcref{tab:supp-time-dependent-auc, tab:supp-time-dependent-auc-human}.

\begin{table}[H]
\centering
\caption{Year-specific time-dependent AUCs in the reading-center-defined analyses.}
\label{tab:supp-time-dependent-auc}
\small
\begin{tabular}{lccc}
\toprule
Cohort
& \makecell[c]{Follow-up horizon, \\years}
& \makecell[c]{Chronological-age model AUC\\(95\% CI)}
& \makecell[c]{FAA model AUC\\(95\% CI)} \\
\midrule
{AREDS}
& 2 & 0.834 (0.646--0.987) & 0.930 (0.870--0.996) \\
& 3 & 0.890 (0.825--0.943) & 0.917 (0.867--0.957) \\
& 4 & 0.880 (0.828--0.926) & 0.899 (0.854--0.937) \\
& 5 & 0.831 (0.760--0.892) & 0.888 (0.837--0.930) \\
& 6 & 0.845 (0.782--0.900) & 0.899 (0.854--0.937) \\
& 7 & 0.858 (0.798--0.910) & 0.908 (0.865--0.943) \\
\midrule
{AREDS2}
& 2 & 0.735 (0.709--0.759) & 0.787 (0.764--0.808) \\
& 3 & 0.739 (0.718--0.759) & 0.795 (0.777--0.812) \\
& 4 & 0.739 (0.720--0.758) & 0.792 (0.775--0.809) \\
& 5 & 0.726 (0.694--0.757) & 0.777 (0.746--0.807) \\
\bottomrule
\end{tabular}
\end{table}

\begin{table}[H]
\centering
\caption{Year-specific time-dependent AUCs in the specialist-defined analyses.}
\label{tab:supp-time-dependent-auc-human}
\small
\begin{tabular}{lccc}
\toprule
Cohort
& \makecell[c]{Follow-up horizon, \\years}
& \makecell[c]{Chronological-age model AUC\\(95\% CI)}
& \makecell[c]{FAA model AUC\\(95\% CI)} \\
\midrule
{AREDS}
& 2 & 0.845 (0.686--0.990) & 0.927 (0.830--0.982) \\
& 3 & 0.834 (0.767--0.894) & 0.876 (0.811--0.933) \\
& 4 & 0.801 (0.728--0.863) & 0.847 (0.787--0.902) \\
& 5 & 0.815 (0.758--0.865) & 0.861 (0.810--0.906) \\
& 6 & 0.817 (0.757--0.870) & 0.864 (0.814--0.909) \\
& 7 & 0.817 (0.761--0.867) & 0.868 (0.820--0.911) \\
\midrule
{AREDS2}
& 2 & 0.636 (0.610--0.661) & 0.710 (0.687--0.733) \\
& 3 & 0.654 (0.634--0.675) & 0.730 (0.712--0.748) \\
& 4 & 0.665 (0.646--0.685) & 0.730 (0.712--0.747) \\
& 5 & 0.652 (0.617--0.686) & 0.720 (0.687--0.752) \\
\bottomrule
\end{tabular}
\end{table}

\newpage

\section{Sample Sizes and Event Counts for Kaplan--Meier Analyses}

\begin{table}[H]
\centering
\caption{Sample sizes and late-AMD progression events in the Kaplan--Meier analyses of the AREDS internal-test cohort. Events indicate eye-level progression to late AMD during follow-up.}
\label{tab:supp-risk-events}
\small
\begin{tabularx}{\linewidth}{
    L
    S[table-format=3]
    S[table-format=3]
    S[table-format=3]
    S[table-format=2]
    S[table-format=2]
}
\toprule
& \multicolumn{3}{c}{Eyes} & \multicolumn{2}{c}{Events} \\
\cmidrule(r){2-4}\cmidrule{5-6}
Model & {Eligible} & {High-risk} & {Low-risk} & {High-risk} & {Low-risk} \\
\midrule
\showrowcolors
Reading center-based Cox model with chronological age
(\zcref{fig:km}a)
& 769 & 201 & 568 & 50 & 24 \\\grayline

Reading center-based Cox model with FAA
(\zcref{fig:km}b)
& 769 & 195 & 574 & 60 & 14 \\\grayline

Specialist-based Cox model with chronological age
(\zcref{fig:km}c)
& 811 & 183 & 628 & 48 & 31 \\\grayline

Specialist-based Cox model with FAA
(\zcref{fig:km}d)
& 811 & 175 & 636 & 51 & 28 \\\grayline

Reading center-based Cox model in eligible eyes from participants in the older population (age $\geq 70$ years, \zcref{fig:km others}a)
& 289 & 138 & 151 & 46 & 6 \\\grayline

Reading center-based Cox model in eligible eyes from participants with FAA age gap $\geq 10$ years (\zcref{fig:km others}b)
& 123 & 39 & 84 & 14 & 12 \\\grayline

Reading center-based Cox model in eligible eyes from participants with a smoking history (\zcref{fig:km others}c)
& 443 & 108 & 335 & 37 & 14 \\\grayline

Specialist-based Cox model in eligible eyes from participants in the older population (age $\geq 70$ years, \zcref{fig:km others}d)
& 303 & 137 & 166 & 40 & 14 \\\grayline

Specialist-based Cox model in eligible eyes from participants with FAA age gap $\geq 10$ years (\zcref{fig:km others}e)
& 132 & 45 & 87 & 15 & 13 \\\grayline

Specialist-based Cox model in eligible eyes from participants with a smoking history (\zcref{fig:km others}f)
& 465 & 104 & 361 & 34 & 21 \\\grayline

FAA in eligible eyes with no AMD (\zcref{fig:km strratified}a)
& 326 & 265 & 61 & 1 & 0 \\\grayline

FAA in eligible eyes with early AMD (\zcref{fig:km strratified}b)
& 328 & 178 & 150 & 27 & 2 \\\grayline

FAA in eligible eyes with intermediate AMD (\zcref{fig:km strratified}c)
& 115 & 59 & 56 & 28 & 16 \\
\bottomrule
\end{tabularx}
\end{table}

\end{document}